\documentclass{article}
\usepackage{amsmath,amssymb}
\usepackage[english]{babel}
\usepackage[colorlinks, linkcolor=black, urlcolor=black]{hyperref}
\usepackage[numbers]{natbib}
\usepackage[dvipsnames]{xcolor}

\definecolor{Green}{RGB}{0,189,107}
\usepackage{titlesec}
\usepackage{graphicx}
\usepackage{float}
\usepackage{changepage}
\usepackage{booktabs}
\usepackage{multirow}
\usepackage[top=3cm, bottom=3cm, right=3cm, left=3cm]{geometry}
\usepackage{caption}
\usepackage{subcaption}
\usepackage{tikz}
\usetikzlibrary{positioning, arrows.meta, shapes.geometric, fit, backgrounds, calc}
\usepackage{tgadventor}
\usepackage[T1]{fontenc}

\makeatletter
\renewcommand{\maketitle}{\bgroup\setlength{\parindent}{0pt}
\begin{flushleft}
  \textbf{\LARGE\textcolor{Green}{\fontfamily{qag}\selectfont \@title}}
  
  \vspace{0.3cm}
  {\fontfamily{qag}\selectfont \large \@author}
  \vspace{0.3cm}
\end{flushleft}\egroup
}
\makeatother

\title{Kandinsky S2V 1.0: Audio-Driven Avatar Video Generation}
\makeatother

\title{Kandinsky S2V 1.0: Audio-Driven Avatar Video Generation}
\author{{\color{Green}\textbf{Kandinsky Lab}$^*$} \\
\small $^*$ A detailed list of the contributors can be found at the end of this paper.}

\titleformat{\section}
  {\normalfont\sffamily\Large\bfseries\color{Green}}
  {\thesection}{1em}{}
\titleformat{\subsection}
  {\normalfont\sffamily\large\bfseries\color{Green}}
  {\thesubsection}{1em}{}
\titleformat{\subsubsection}
  {\normalfont\sffamily\normalsize\bfseries\color{Green}}
  {\thesubsubsection}{1em}{}
  
\begin{document}


\maketitle

\begin{figure}[h!]
\centering
\resizebox{\textwidth}{!}{%
\begin{tikzpicture}[
    every node/.style={align=center, inner sep=2pt},
    feat/.style={font=\bfseries\large, text=Green},
    desc/.style={font=\scriptsize, text width=4.5cm, text=black!75},
    photo/.style={inner sep=0pt, draw=gray!50, line width=0.4pt},
    photolbl/.style={font=\scriptsize\bfseries},
    tslbl/.style={font=\scriptsize, text=black!70},
    plus/.style={font=\LARGE\bfseries, text=Green},
]

\node[feat] (f1) at (2.3,  4.7) {Identity-Faithful};
\node[desc, below=0.05 of f1] {Animates a single reference image with lip-sync, head pose, and gesture coherent with speech.};

\node[feat] (f2) at (7.85,  4.7) {Multilingual};
\node[desc, below=0.05 of f2] {Native English and Russian prompts over a multilingual video corpus.};

\node[feat] (f3) at (13.25, 4.7) {Long-form};
\node[desc, below=0.05 of f3] {Autoregressive 5\,s chunks (25 FPS) held together by a FramePack memory.};

\node[feat] (f4) at (18.65, 4.7) {Two Scales};
\node[desc, below=0.05 of f4] {A fast 2B-parameter model and a more advanced 21B variant.};


\node[photo] (ref) at (1.85, 1.7) {\includegraphics[width=3.5cm,height=2.33cm]{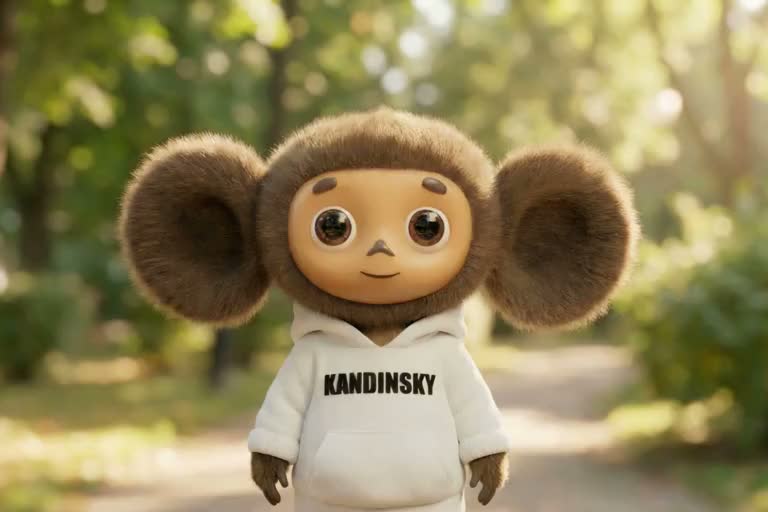}};
\node[photolbl, below=0.08 of ref] {Reference Image};

\node[plus] at (4.85, 1.5) {$+$};

\node (audio) at (4.85, 2.35) {};
\foreach \i/\h in {0/0.12, 1/0.28, 2/0.14, 3/0.36, 4/0.20, 5/0.28, 6/0.14, 7/0.32, 8/0.18, 9/0.12}
  \draw[line width=1pt, Green] (4.40 + \i*0.10, 2.35 - \h/2) -- (4.40 + \i*0.10, 2.35 + \h/2);
\node[photolbl] at (4.85, 1.95) {Audio};

\node[photolbl] (txtdesc) at (4.85, 1.05) {Text Description};

\draw[gray!40, line width=0.4pt] (5.95, 0.4) -- (5.95, 3.3);

\node[photo] (s0) at (7.85, 1.7) {\includegraphics[width=3.5cm,height=2.33cm]{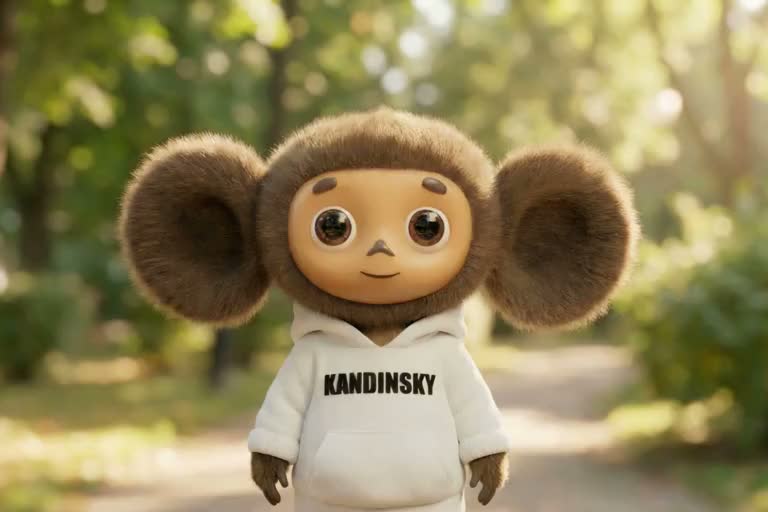}};
\node[photo] (s1) at (11.45, 1.7) {\includegraphics[width=3.5cm,height=2.33cm]{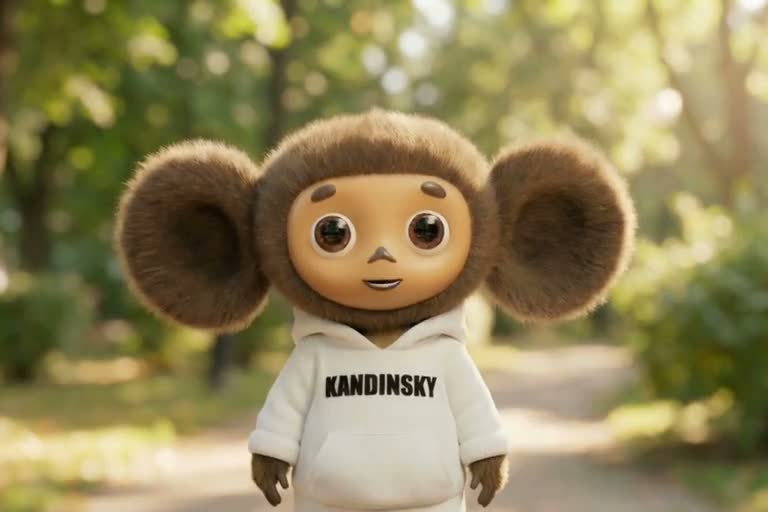}};
\node[photo] (s2) at (15.05, 1.7) {\includegraphics[width=3.5cm,height=2.33cm]{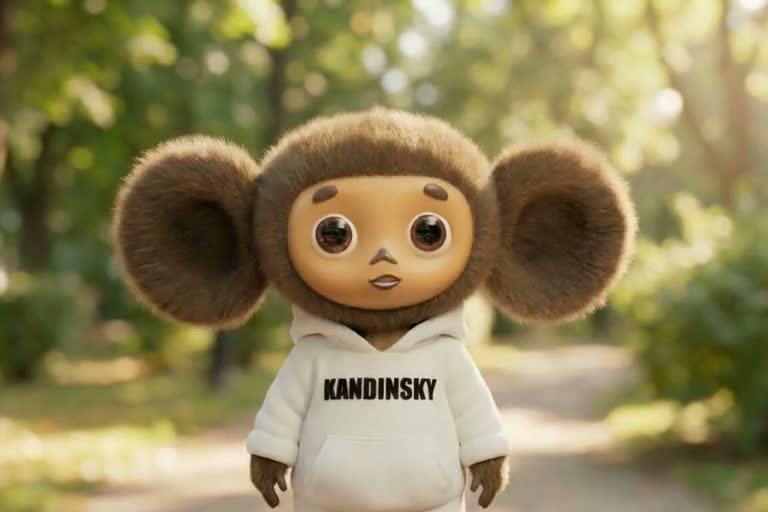}};
\node[photo] (s3) at (18.65, 1.7) {\includegraphics[width=3.5cm,height=2.33cm]{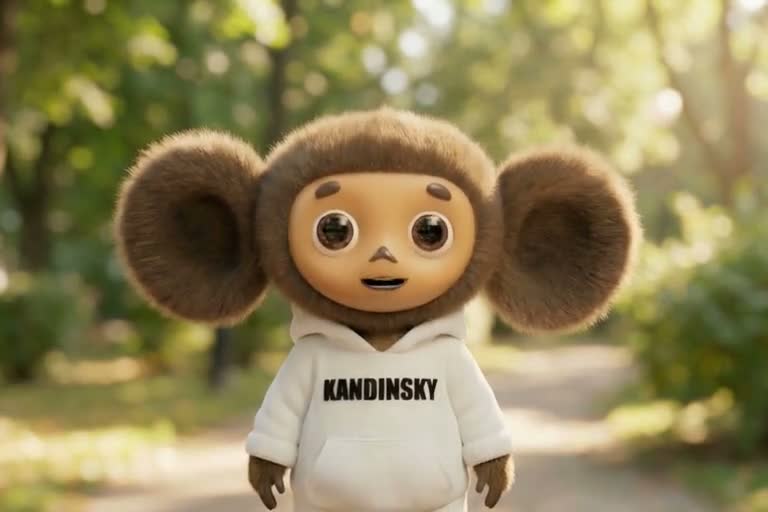}};

\node[tslbl, below=0.08 of s0] {$0$\,s};
\node[tslbl, below=0.08 of s1] {$5$\,s};
\node[tslbl, below=0.08 of s2] {$10$\,s};
\node[tslbl, below=0.08 of s3] {$15$\,s};

\end{tikzpicture}%
}
\end{figure}

\noindent
\textbf{\textcolor{Green}{Abstract:}}
We present Kandinsky S2V 1.0, an open family of speech-to-video models for controllable audio-driven human animation, comprising Kandinsky S2V 1.0 Lite (2B parameters) and Kandinsky S2V 1.0 Pro (21B parameters). From a single reference image, a speech waveform, and a text prompt, the models generate long-form 25 FPS video in which lip motion, expression, pose, and gesture follow the speech. The architecture draws text and audio conditioning from one Qwen2.5-Omni model and adds a reference-aware three-scale visual memory. Training starts from the Kandinsky 5.0 Video foundation models \cite{kandinskylab2025kandinsky5} and combines mixed reference-memory layouts with an area-adaptive face loss. The training corpus contains $1.17$M single-speaker clips selected by active-speaker detection, with captions generated directly in English and Russian. A side-by-side human evaluation against the strongest open-source baseline and two proprietary commercial models shows that Kandinsky S2V 1.0 Pro performs comparably to or better than these models on average and is also preferred for lip synchronization. On common speech to video benchmarks, Kandinsky S2V 1.0 Lite and Kandinsky S2V 1.0 Pro rank among the top performing models according to mean rank over five reported metrics. Beyond the base models, we provide $\pi$-Flow distilled variants of both Kandinsky S2V 1.0 Lite and Kandinsky S2V 1.0 Pro. The distilled Kandinsky S2V 1.0 Pro model generates video in 16 sampling steps without classifier-free guidance, achieving a $5.4\times$ end-to-end speedup over the base model, while matching or exceeding it on nearly all reported benchmark metrics. To accelerate open research and deployment in speech-to-video generation, we release the code, model checkpoints, and diffusers integration under the MIT license.

\noindent
\textbf{\textcolor{Green}{Date:}} October 15, 2026


\noindent
\textbf{\textcolor{Green}{Project Page: }} \href{https://kandinskyavatar.github.io/}{https://kandinskyavatar.github.io/}


\noindent
\textbf{\textcolor{Green}{GitHub:}} \href{https://github.com/kandinskylab/kandinsky-s2v}{https://github.com/kandinskylab/kandinsky-s2v}

\noindent
\textbf{\textcolor{Green}{Hugging Face:}} \href{https://huggingface.co/collections/kandinskylab/kandinsky-s2v-10}{https://huggingface.co/collections/kandinskylab/kandinsky-s2v-10}


\newpage
\tableofcontents

\newpage

\section[Introduction]{\color{Green}Introduction}\label{sec:introduction}

Audio-driven human animation aims to turn a single reference image and a driving speech signal into a video in which lip motions, facial expressions, head pose, and body movements are coherent with the audio. Recent approaches build on large-scale text-to-video generation \cite{kong2025hunyuanvideosystematicframeworklarge, wan2025wan} and add two constraints: preserving the identity of the person in the reference image and aligning motion with the audio. Applications include localized dubbing and conversational digital humans.

Until recently this was largely the domain of two-stage pipelines, in which an audio-to-motion model produced intermediate representations such as 3DMM parameters or facial landmarks and a separate renderer or diffusion model generated video from these representations \cite{zhang2023sadtalker, wei2024aniportrait}. Direct audio-to-video methods extended image diffusion U-Nets with temporal layers and demonstrated strong lip-sync on short head-only clips \cite{xu2024hallo, tian2024emo}. Recent systems build on pretrained video Diffusion Transformers and generate half-body or full-body motion \cite{chen2025hunyuanvideo, kong2025multitalk, yang2025infinitetalk, gan2025omniavatar, humanaigc2025wans2v, longcatteam2026longcatavatar15}.

Kandinsky S2V 1.0 also uses a Diffusion Transformer architecture. The family comprises two controllable speech-to-video models with 2B and 21B parameters, both publicly released. Both models accept text prompts in English and Russian and are trained on a multilingual video corpus with substantial Russian-language content. They generate long videos by composing 5-second chunks at 25 FPS with an explicit visual memory.

\subsection{Recent Open-Source S2V Models}
We briefly review recent open-source S2V systems that serve as baselines in our comparisons.

\paragraph{HunyuanVideo-Avatar \cite{chen2025hunyuanvideo}.}
It is built on a multimodal diffusion transformer and targets dynamic single- and multi-character animation. It replaces addition-based character conditioning with a dedicated character image injection module. A separate emotion reference image provides expression cues, while face-aware audio adapters use latent face masks to route each speech stream to the corresponding character.

\paragraph{MultiTalk \cite{kong2025multitalk}.}
It introduces audio-driven multi-person conversational video generation on the Wan2.1-I2V-14B backbone. Naive multi-stream audio injection can associate a voice with the wrong speaker. Label Rotary Position Embedding addresses this problem by assigning matched labels to audio embeddings and person regions in the video latents, encouraging audio cross-attention to focus on the corresponding speaker regions. Partial-parameter and multi-task training preserve the instruction-following ability of the base model.

\paragraph{InfiniteTalk \cite{yang2025infinitetalk}.}
It builds on MeiGen-MultiTalk and extends it to sparse-frame video dubbing and image-to-video animation. Given a source video or a reference image and a new audio track, it generates audio-driven full-body motion. In video dubbing, sparse keyframe conditioning helps preserve identity, background, and the overall camera trajectory. Temporal context frames smooth transitions between chunks, and a reference-frame sampling strategy balances control strength during streaming long-form generation.

\paragraph{OmniAvatar \cite{gan2025omniavatar}.}
It is built on the Wan2.1-T2V-14B foundation. It injects pixel-wise multi-hierarchical audio features directly into the latent video representation instead of relying on audio cross-attention. LoRA adaptation preserves the text-to-video prior of the base model, while the audio conditioning extends motion control beyond the face to adaptive full-body animation.

\paragraph{Wan-S2V \cite{humanaigc2025wans2v}.}
It also builds on a Wan-14B backbone, but targets cinematic film and television scenes rather than isolated talking heads. It encodes speech with Wav2Vec, inserts Audio Blocks that compute frame-wise attention between visual and audio tokens, and applies FramePack \cite{zhang2025frame} to compress older motion-frame tokens for long-form continuation.

\paragraph{StableAvatar \cite{tu2025stableavatar}.}
It generates long audio-driven videos from a reference image and speech with a video diffusion transformer. A time-step-aware audio adapter limits error accumulation across chunks. At inference, audio-native guidance and a dynamic weighted sliding window support speech alignment and smooth transitions.

\paragraph{LongCat 1.5 \cite{longcatteam2025longcat, longcatteam2026longcatavatar15}.}
It is built on the LongCat-Video foundation, a UMT5-conditioned DiT, and supports audio-text-to-video, audio-text-image-to-video, and audio-driven continuation in one model. Disentangled Unconditional Guidance distinguishes silent-audio embeddings from null audio conditions, helping maintain natural motion during pauses in speech. Reference Skip Attention, applied at inference, reduces copying of the reference pose. Cross-chunk latent stitching avoids repeated VAE decoding and encoding between chunks. Version 1.5, which we compare against, replaces the Wav2Vec audio encoder with Whisper Large and uses step distillation to enable inference with $8$ function evaluations.

\subsection{Kandinsky S2V 1.0}
Kandinsky S2V 1.0 is built on the Kandinsky 5.0 Video foundation models \cite{kandinskylab2025kandinsky5}. The reference image anchors identity, appearance, and scene content. Speech drives lip motion and other speech-related movement. The text prompt describes how the avatar speaks and the emotions it expresses.

The main design choices behind Kandinsky S2V 1.0 can be grouped into three components.

\begin{itemize}
  \item {\bfseries Architecture.} Most recent S2V systems use separate pretrained models to extract text and audio features. MultiTalk, InfiniteTalk, OmniAvatar, and Wan-S2V use Wav2Vec or Wav2Vec2 features \cite{kong2025multitalk, yang2025infinitetalk, gan2025omniavatar, humanaigc2025wans2v}, while HunyuanVideo-Avatar and LongCat 1.5 use Whisper \cite{chen2025hunyuanvideo, longcatteam2026longcatavatar15}. Kandinsky S2V 1.0 extracts text and audio features from the Thinker and audio tower of a single Qwen2.5-Omni model respectively \cite{xu2025qwen25omni}. Trainable text and audio encoders in the DiT adapt the Qwen2.5-Omni features before they enter separate conditioning paths. We also adapt FramePack \cite{zhang2025frame} into a reference-aware three-scale visual memory for long-form video generation. We keep the identity reference separate from the motion history, with its own segment embedding and its own temporal position.
  \item {\bfseries Training.} Mixed reference-memory layouts outside the target interval on the temporal RoPE axis, either near the target or at the beginning of a longer context, discouraging reference-pose copying at chunk boundaries. An area-normalized face loss maintains the contribution of small facial regions to the training objective.
  \item {\bfseries Data.} We assemble a corpus of $1.17$M single-speaker clips from the Kandinsky 5.0 video corpus \cite{kandinskylab2025kandinsky5} and the Panda-70M portion of TalkVerse \cite{wang2025talkverse, chen2024panda70m}. We use LR-ASD \cite{liao2025lrasd} to retain clips with one uninterrupted active-speaker track, with technical-quality filtering applied to the Kandinsky portion and deduplication to the whole corpus. Qwen3-VL \cite{bai2025qwen3vltechnicalreport} annotates each clip at three levels of granularity, using separate prompts for English and Russian captions. The final corpus contains $53.9$\% English and $25.8$\% Russian speech, supporting Russian-language text conditioning without a separate fine-tuning stage.
\end{itemize}

\subsection{Contributions}
Our main contributions are as follows:

\begin{itemize}
  \item We introduce and release two Kandinsky S2V 1.0 models, Kandinsky S2V 1.0 Lite (2B) and Kandinsky S2V 1.0 Pro (21B), that can generate SD and HD video at 25 FPS with multilingual prompting and autoregressive long-form continuation.
  \item We develop an architecture and training recipe that combine text and audio conditioning from a single Qwen2.5-Omni model, mixed reference-memory layouts, an area-normalized face loss, and a curated multilingual video corpus.
  \item A side-by-side human evaluation against the strongest open-source baseline and two proprietary commercial models shows that Kandinsky S2V 1.0 Pro performs comparably to or better than these models on average and is also preferred for lip synchronization.
  \item On EMTD, CelebV-HQ, and HDTF, distilled Kandinsky S2V 1.0 Pro achieves the best mean rank and min-max score among all evaluated models, with both measures aggregating the five reported metrics.
  \item We provide $\pi$-Flow distilled variants of both Kandinsky S2V 1.0 Lite and Kandinsky S2V 1.0 Pro. The distilled Pro model generates video in 16 sampling steps without classifier-free guidance, achieving a $5.4\times$ end-to-end speedup over the base Pro model while matching or exceeding it on nearly all reported benchmark metrics.
  \item We release all model weights, source code, and integration with the diffusers library to the research community under the MIT license.
\end{itemize}

The remainder of this report is organized as follows. Section~\ref{sec:data_processing_pipeline} describes the data collection, filtering, and captioning pipeline that produces the training corpus. Section~\ref{sec:k5_s2v_framework} presents the model architecture. Section~\ref{sec:training_stages} describes continuous pre-training and few-step distillation. Section~\ref{subsec:long_form_inference} explains long-form inference. Section~\ref{sec:evaluation} compares the base and distilled models with competing systems. Section~\ref{sec:conclusion} summarizes the contributions.
\section[Data Processing Pipeline]{\color{Green}Data Processing Pipeline}\label{sec:data_processing_pipeline}

Unlike generic text-to-video data, training data for audio-driven human animation must maintain a clear association between speech and the same visible speaker throughout each clip. Active-speaker detection is therefore the main filter of our pipeline, followed by additional technical-quality filtering. An overview of the pipeline is provided in Table~\ref{tab:data_pipeline}.

\subsection{Data Sources}

The first source is the Kandinsky 5.0 video corpus \cite{kandinskylab2025kandinsky5}, whose scenes are sourced from various open datasets and large video platforms and segmented by shot detection. We refer to it below as the web collection. It includes the Russian-language content that motivates the bilingual conditioning of Section~\ref{sec:k5_s2v_framework}. Section~\ref{subsec:provenance} describes its provenance and the procedure for excluding evaluation data. We retain the scenes that contain at least one detected face and last longer than $10$ seconds, which gives a base set of approximately $3.8$M videos. This set is then annotated and filtered by active-speaker detection (Section~\ref{subsec:asd}), followed by technical-quality filtering. Blurred lips and hands reduce the quality of supervision for lip motion and gestures, respectively. We therefore score the surviving videos with Q-Align \cite{wu2023qalign} in its technical-quality mode and keep those with scores above $4.0$.

The second source is the public TalkVerse corpus \cite{wang2025talkverse}: we use its Panda-70M \cite{chen2024panda70m} portion and keep the videos longer than $5$ seconds, which gives $700$K videos. It is already filtered for technical quality, so no further quality stage is applied.

\subsection{Active-Speaker Detection and Clip Extraction}\label{subsec:asd}

Both sources pass through the same active-speaker stage. We run LR-ASD \cite{liao2025lrasd},\footnote{We replace the default face detector of LR-ASD with YOLOv11n-face \cite{yoloface}, a lighter and faster detector that also improves detection quality on our data.} which yields face tracks with per-frame ASD scores indicating whether the tracked person is speaking. All thresholds below are applied to this score after smoothing with a moving average over $5$ frames. A clip is used only if a single face track covers it without interruption. Since tracks can be interrupted by  fast head turns, occlusions, and cuts, this acts as an implicit filter on face motion.

From the web collection we cut fixed-length $10$-second clips. Within each uninterrupted face track we average the smoothed ASD-score over every $10$-second window and take the arg-max window, keeping it only if it contains exactly one detected active speaker and its average exceeds $2.0$. The window search recovers usable material from videos that are unusable as a whole and favors windows with stronger average evidence of correspondence between speech and the visible face.

For the TalkVerse subset we threshold the maximum of the smoothed ASD-score rather than its mean, using this as an additional check for active speech. Exactly one clip is taken per video, always from its beginning, with its length set by the duration of the source: a video shorter than $10$ seconds yields a $5$-second clip, a video of $10$ seconds or longer a $10$-second one. The clip is kept only if the face track is unbroken over that span. There is no window search, since the subset is already filtered for lip-sync. Selecting the opening segment avoids explicitly optimizing the window position for the ASD score. The resulting clips begin at the source scene boundaries.
The two clip lengths correspond to the two conditioning regimes of Section~\ref{subsubsec:ref_memory_placement}. A $5$-second clip is exactly one generation chunk of $125$ frames at $25$ FPS and is used for the training steps without memory conditioning. A $10$-second clip is split into two halves: the first becomes the memory clip $V_{\text{mem}}$, the second the chunk the model is asked to denoise. Only $10$-second clips therefore supply memory-conditioned steps, which is why the web collection is cut at that length throughout.

The corpus is processed jointly across both sources to remove duplicates: we embed the source videos with InternVideo2-1B \cite{wang2024internvideo2}, cluster the embeddings, and subsample the anomalously large clusters, which brings the web part to $610$K clips and the TalkVerse part to $563$K.

\subsection{Captioning}\label{subsec:captioning}

Every retained clip is annotated by Qwen3-VL-32B-Instruct \cite{bai2025qwen3vltechnicalreport}, served through vLLM \cite{kwon2023vllm} with FP8 weights and an FP8 key-value cache. Frames are sampled at $2$ FPS and resized to an area of at most $512 \times 512$ pixels.

For a $10$-second clip only the second half is captioned, that is the last $5$ seconds, giving $10$ frames: the first half serves as memory conditioning and is not a denoising target in this training example, so the caption must describe the target chunk alone, as it does at inference. Five-second clips are captioned in full.

The system prompt is written for audio-driven human animation rather than generic video description: it asks for lip dynamics in articulatory terms, facial micro-expressions described through visible muscle changes, the rhythm of visible speech and its coordination with head motion, body posture and motion quality, camera movement, and the static content of the scene. It forbids inventing spoken content, to reduce the risk of transcript-like content entering the text condition. It also forbids timestamps and frame indices to avoid references to specific positions within the source video.

Three levels of granularity are produced in a single pass: an exhaustive description of $300$ or more words, a standard caption of $150$ to $200$ words, and a summary of $30$ to $50$ words. Two such requests are issued per clip, one constrained to English output and one to Russian, so the Russian captions are generated natively rather than translated. The resulting six captions per clip are sampled at training time with fixed per-language, per-granularity probabilities.

\subsection{Provenance and Exclusion of Evaluation Data}\label{subsec:provenance}

Both parts of the corpus consist of publicly posted video of real people, as do the datasets on which audio-driven animation is commonly trained and evaluated, such as VoxCeleb, HDTF, CelebV-HQ, and TalkVerse. The web collection is the subset of the Kandinsky 5.0 video corpus \cite{kandinskylab2025kandinsky5} that survives the pipeline above and inherits the provenance documented there, scenes sourced from open datasets and large video platforms. The TalkVerse part comes from a public dataset that is distributed as metadata, source URLs with timestamps and quality scores, rather than as video. The corpus itself, that is the source videos, the extracted clips, their captions, and the lists of source URLs, is not distributed.

Evaluation data are excluded from the training corpus at two levels, and the check runs over both parts of the corpus before captioning. At the source level, training videos that share a source video identifier with a benchmark clip are dropped. At the identity level, ArcFace embeddings \cite{deng2019arcface} of the benchmark faces are compared with the face tracks of the training clips, and clips that match a benchmark speaker above the same-identity threshold are removed, so the reported scores measure generalization to unseen people rather than to unseen clips of seen people. The reference images and audio of the side-by-side sets of Section~\ref{sec:human_eval} are collected outside the training pipeline and pass the same two checks.

\subsection{Resulting Corpus}

The final corpus consists of $1.17$M clips, split almost evenly between the two sources: the web part supplies the Russian-language coverage, the TalkVerse part high-resolution single-speaker footage. Figure~\ref{fig:lang_mix} shows the distribution of spoken languages: English accounts for slightly over half of the corpus and Russian for roughly a quarter. Training with the bilingual captions supports Russian-language text conditioning without a dedicated language-specific fine-tuning stage.

\begin{table}[H]
\centering
\caption{Data pipeline and the resulting corpus. Q-Align technical filtering applies only to the web collection. The active-speaker stage and the exclusion of evaluation data are shared, and deduplication runs over the union of the two sources.}
\label{tab:data_pipeline}
\footnotesize
\begin{tabular}{lcc}
\toprule
Stage & \shortstack{Web\\collection} & \shortstack{TalkVerse\\(Panda-70M part)} \\
\midrule
Videos with a detected face, after duration filtering & $> 10$\,s: $3.8$M & $> 5$\,s: $700$K \\
Single-speaker, smoothed ASD-score $> 2.0$ & window mean & max \\
Q-Align technical score $> 4.0$ & \checkmark & Not applied \\
Cluster-based deduplication & \multicolumn{2}{c}{\checkmark, over the whole corpus} \\
Exclusion of evaluation data (Section~\ref{subsec:provenance}) & \checkmark & \checkmark \\
\midrule
Clip length used for training & $10$\,s & $5$\,s and $10$\,s \\
Retained clips & $610$K & $563$K \\
\midrule
Total & \multicolumn{2}{c}{$1.17$M clips} \\
\bottomrule
\end{tabular}
\end{table}

\definecolor{pieEN}{HTML}{4A6E8A}
\definecolor{pieKK}{HTML}{E8A33D}
\definecolor{piePT}{HTML}{5FA85F}
\definecolor{pieES}{HTML}{D97B66}
\definecolor{pieFR}{HTML}{3FA8A0}
\definecolor{pieOT}{HTML}{B0B0B0}
\colorlet{pieRU}{Green}

\begin{figure}[H]
\centering
\begin{tikzpicture}[scale=0.85, transform shape]
\filldraw[pieEN, draw=white, line width=0.7pt] (0,0) -- (90.00:1.75) arc[start angle=90.00, end angle=-103.90, radius=1.75] -- cycle;
\filldraw[pieRU, draw=white, line width=0.7pt] (0,0) -- (-103.90:1.75) arc[start angle=-103.90, end angle=-196.66, radius=1.75] -- cycle;
\filldraw[pieKK, draw=white, line width=0.7pt] (0,0) -- (-196.66:1.75) arc[start angle=-196.66, end angle=-215.38, radius=1.75] -- cycle;
\filldraw[piePT, draw=white, line width=0.7pt] (0,0) -- (-215.38:1.75) arc[start angle=-215.38, end angle=-227.16, radius=1.75] -- cycle;
\filldraw[pieES, draw=white, line width=0.7pt] (0,0) -- (-227.16:1.75) arc[start angle=-227.16, end angle=-233.40, radius=1.75] -- cycle;
\filldraw[pieFR, draw=white, line width=0.7pt] (0,0) -- (-233.40:1.75) arc[start angle=-233.40, end angle=-239.49, radius=1.75] -- cycle;
\filldraw[pieOT, draw=white, line width=0.7pt] (0,0) -- (-239.49:1.75) arc[start angle=-239.49, end angle=-270.00, radius=1.75] -- cycle;

\begin{scope}[shift={(2.45,1.26)}]
\fill[pieEN] (0,0.00) rectangle ++(0.26,0.26);
\node[anchor=west, font=\small] at (0.40,0.13) {English};
\node[anchor=east, font=\small] at (4.4,0.13) {$53.9$\%};
\fill[pieRU] (0,-0.42) rectangle ++(0.26,0.26);
\node[anchor=west, font=\small] at (0.40,-0.29) {Russian};
\node[anchor=east, font=\small] at (4.4,-0.29) {$25.8$\%};
\fill[pieKK] (0,-0.84) rectangle ++(0.26,0.26);
\node[anchor=west, font=\small] at (0.40,-0.71) {Kazakh};
\node[anchor=east, font=\small] at (4.4,-0.71) {$5.2$\%};
\fill[piePT] (0,-1.26) rectangle ++(0.26,0.26);
\node[anchor=west, font=\small] at (0.40,-1.13) {Portuguese};
\node[anchor=east, font=\small] at (4.4,-1.13) {$3.3$\%};
\fill[pieES] (0,-1.68) rectangle ++(0.26,0.26);
\node[anchor=west, font=\small] at (0.40,-1.55) {Spanish};
\node[anchor=east, font=\small] at (4.4,-1.55) {$1.7$\%};
\fill[pieFR] (0,-2.10) rectangle ++(0.26,0.26);
\node[anchor=west, font=\small] at (0.40,-1.97) {French};
\node[anchor=east, font=\small] at (4.4,-1.97) {$1.7$\%};
\fill[pieOT] (0,-2.52) rectangle ++(0.26,0.26);
\node[anchor=west, font=\small] at (0.40,-2.39) {Other languages};
\node[anchor=east, font=\small] at (4.4,-2.39) {$8.5$\%};
\end{scope}
\end{tikzpicture}
\caption{\small Distribution of spoken languages over the $1.17$M training clips. Languages below $1.5$\% are pooled into Other languages.}
\label{fig:lang_mix}
\end{figure}
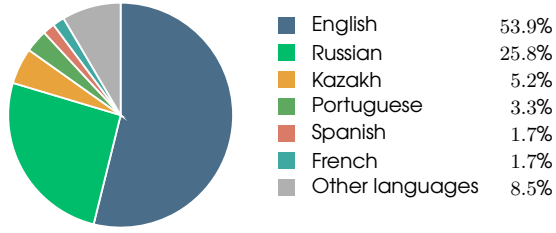
\section{Kandinsky S2V 1.0 Architecture}\label{sec:k5_s2v_framework}

Kandinsky S2V 1.0 turns a single reference image into a talking avatar video driven by speech, with a text prompt describing how the avatar speaks. It is built on the Kandinsky 5.0 Video foundation models \cite{kandinskylab2025kandinsky5}. Like its base model, Kandinsky S2V 1.0 is a 3D Diffusion Transformer that operates in the latent space of a pretrained video VAE \cite{kong2025hunyuanvideosystematicframeworklarge} and is trained with a flow-matching objective \cite{lipman2023flow, liu2022flowstraightfastlearning, esser2024scalingrectifiedflowtransformers}. The model generates video in $5$-second chunks of $125$ frames at $25$ FPS. Each chunk is conditioned on the reference image, the speech, and the prompt, and every chunk after the first also receives a visual memory of the frames generated before it. Kandinsky S2V 1.0 comes in two sizes, Lite (2B) and Pro (21B), and we release both. They differ in visual depth, model and feed-forward dimensions, the depth of the text and audio stacks, head dimension, and the audio injection schedule.

\begin{figure}[H]
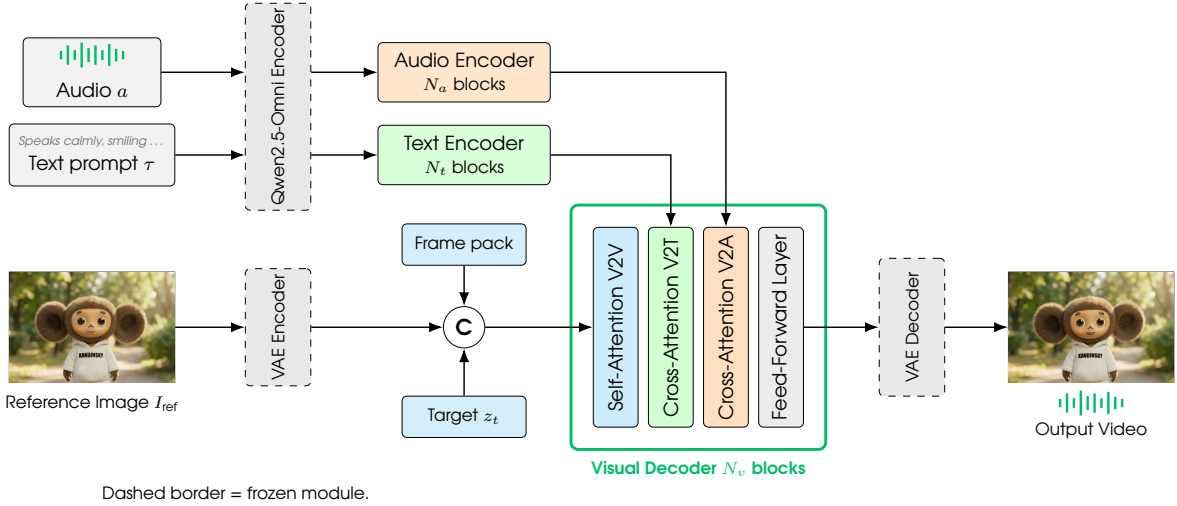

\centering
\resizebox{\textwidth}{!}{%
\begin{tikzpicture}[
    font=\footnotesize,
    every node/.style={align=center, inner sep=4pt},
    inp/.style={rectangle, rounded corners=2pt, draw, fill=gray!10,
                minimum height=0.7cm, minimum width=1.9cm},
    inpphoto/.style={rectangle, sharp corners, draw, fill=gray!12,
                     minimum height=1.0cm, minimum width=1.9cm},
    frozen/.style={rectangle, rounded corners=2pt, draw, dashed, fill=gray!18,
                   minimum width=0.95cm, font=\scriptsize, inner sep=2pt},
    txtenc/.style={rectangle, rounded corners=2pt, draw, fill=green!15,
                   minimum height=0.85cm, minimum width=2.5cm},
    audenc/.style={rectangle, rounded corners=2pt, draw, fill=orange!20,
                   minimum height=0.85cm, minimum width=2.5cm},
    framepack/.style={rectangle, rounded corners=2pt, draw, fill=cyan!15,
                      minimum height=0.6cm, minimum width=1.8cm, font=\scriptsize},
    target/.style={rectangle, rounded corners=2pt, draw, fill=cyan!15,
                   minimum height=0.6cm, minimum width=1.8cm, font=\scriptsize},
    cnode/.style={circle, draw, fill=white, minimum size=0.6cm, inner sep=0, font=\bfseries\small},
    attnV/.style={rectangle, rounded corners=2pt, draw, fill=cyan!18,
                  minimum height=2.9cm, minimum width=0.65cm, inner sep=1pt},
    attnT/.style={rectangle, rounded corners=2pt, draw, fill=green!18,
                  minimum height=2.9cm, minimum width=0.65cm, inner sep=1pt},
    attnA/.style={rectangle, rounded corners=2pt, draw, fill=orange!25,
                  minimum height=2.9cm, minimum width=0.65cm, inner sep=1pt},
    ffnblk/.style={rectangle, rounded corners=2pt, draw, fill=gray!15,
                  minimum height=2.9cm, minimum width=0.65cm, inner sep=1pt},
    arr/.style={-{Latex[length=2mm, width=1.5mm]}, semithick},
    arrd/.style={-{Latex[length=2mm, width=1.5mm]}, semithick, dashed},
    note/.style={rectangle, rounded corners=2pt, draw, dashed, fill=gray!10,
                 inner sep=4pt, font=\scriptsize, align=center},
]

\node[inp, minimum height=1.0cm, minimum width=2.0cm, inner sep=0pt] (aud) at (0, 3.5) {};
\foreach \i/\h in {0/0.12, 1/0.28, 2/0.14, 3/0.36, 4/0.20, 5/0.28, 6/0.14, 7/0.32, 8/0.18, 9/0.12}
  \draw[line width=1pt, Green] (-0.45 + \i*0.10, 3.75 - \h/2) -- (-0.45 + \i*0.10, 3.75 + \h/2);
\node[font=\footnotesize] at (0, 3.25) {Audio $a$};

\node[inp, minimum height=0.9cm, minimum width=2.4cm] (txt) at (0, 2.3)
    {\textit{\tiny\color{black!55}Speaks calmly, smiling \dots}\\[1pt]Text prompt $\tau$};
\node[inpphoto, minimum height=1.6cm, minimum width=2.4cm, inner sep=0pt] (ref) at (0, -0.2) {\includegraphics[width=2.4cm,height=1.6cm]{frames/ref.jpg}};
\node[font=\scriptsize] at (0, -1.3) {Reference Image $I_\text{ref}$};

\node[frozen, minimum height=3.0cm] (qwen) at (2.7, 3.0)
    {\rotatebox{90}{Qwen2.5-Omni Encoder}};
\node[frozen, minimum height=1.7cm] (vae)  at (2.7, -0.2)
    {\rotatebox{90}{VAE Encoder}};

\node[audenc] (audE) at (5.4, 3.5) {Audio Encoder\\\scriptsize $N_a$ blocks};
\node[txtenc] (txtE) at (5.4, 2.3) {Text Encoder\\\scriptsize $N_t$ blocks};

\node[framepack] (fp)     at (5.4, 1.0)  {Frame pack};
\node[cnode]     (concat) at (5.4, -0.2) {C};
\node[target]    (tgt)    at (5.4, -1.5) {Target $z_t$};

\node[attnV]  (sa)  at (7.6, -0.2)  {\rotatebox{90}{Self-Attention V2V}};
\node[attnT]  (cat) at (8.4, -0.2)  {\rotatebox{90}{Cross-Attention V2T}};
\node[attnA]  (caa) at (9.2, -0.2)  {\rotatebox{90}{Cross-Attention V2A}};
\node[ffnblk] (ffn) at (10.0, -0.2) {\rotatebox{90}{Feed-Forward Layer}};

\node[draw=Green, very thick, rounded corners=3pt,
      fit=(sa)(cat)(caa)(ffn), inner sep=9pt,
      label={[font=\scriptsize\bfseries, anchor=north, text=Green]south:Visual Decoder $N_v$ blocks}]
      (dit) {};

\node[frozen, minimum height=2.0cm] (vaed) at (11.9, -0.2)
    {\rotatebox{90}{VAE Decoder}};

\node[inpphoto, minimum height=1.6cm, minimum width=2.4cm, inner sep=0pt] (outvid)
    at (14.5, -0.2) {\includegraphics[width=2.4cm,height=1.6cm]{frames/f1.jpg}};
\foreach \i/\h in {0/0.12, 1/0.28, 2/0.14, 3/0.36, 4/0.20, 5/0.28, 6/0.14, 7/0.32, 8/0.18, 9/0.12}
  \draw[line width=1pt, Green] (14.5 - 0.45 + \i*0.10, -1.3 - \h/2) -- (14.5 - 0.45 + \i*0.10, -1.3 + \h/2);
\node[font=\scriptsize] at (14.5, -1.7) {Output Video};

\draw[arr] (aud.east) -- (qwen.west |- aud);
\draw[arr] (txt.east) -- (qwen.west |- txt);
\draw[arr] (ref.east) -- (vae.west);

\draw[arr] (qwen.east |- aud) -- (audE.west);
\draw[arr] (qwen.east |- txt) -- (txtE.west);

\draw[arr] (vae.east) -- (concat.west);

\draw[arr] (fp.south) -- (concat.north);
\draw[arr] (tgt.north) -- (concat.south);

\draw[arr] (audE.east) -| (caa.north);
\draw[arr] (txtE.east) -| (cat.north);

\draw[arr] (concat.east) -- (sa.west);

\draw[arr] (ffn.east) -- (vaed.west);

\draw[arr] (vaed.east) -- (outvid.west);

\node[font=\scriptsize, anchor=west] at (0, -2.6) {Dashed border = frozen module.};

\end{tikzpicture}%
}
\caption{Overview of the Kandinsky S2V 1.0 architecture. Text features from the frozen Qwen2.5-Omni Thinker and audio features from its frozen audio tower \cite{xu2025qwen25omni} feed trainable text and audio encoder stacks in the DiT, with $N_t$ text blocks and $N_a$ audio blocks. The reference image is encoded by a frozen Hunyuan 3D VAE \cite{kong2025hunyuanvideosystematicframeworklarge}, and its latent is concatenated (denoted by C) with the noisy target latent $z_t$ and a multi-scale FramePack \cite{zhang2025frame} memory of past frames to form the visual sequence consumed by the DiT. Each DiT block applies self-attention (V2V), cross-attention to the text tokens (V2T), an optional cross-attention to the audio tokens (V2A) injected periodically across the stack, and a feed-forward layer. The denoised visual latent is decoded back to pixel space by the frozen VAE decoder.}
\label{fig:k5_s2v_overview}
\end{figure}

\subsection{Conditioning streams}

Figure~\ref{fig:k5_s2v_overview} gives an overview of the architecture. The model takes four inputs from three modalities: a text prompt, a speech waveform, a reference image, and a compressed memory of past frames.

\begin{itemize}
  \item A textual prompt $\tau$ that describes the avatar's manner of speaking and expressed emotions. The frozen Qwen2.5-Omni Thinker \cite{xu2025qwen25omni} encodes it into a sequence of token features of dimension $3584$. The prompt is placed before the audio segment in the Thinker input. Since the Thinker uses causal attention, the extracted prompt features do not depend on the later audio tokens. A trainable text encoder in the DiT adapts these features before the V2T conditioning path. A pooled CLIP \cite{radford2021clip} sentence embedding of dimension $768$ is used as a global style vector that is fused with the diffusion time embedding.
  \item A speech waveform $a$ sampled at $16$ kHz. The Qwen2.5-Omni audio tower encodes it into an audio token stream of dimension $3584$. Kandinsky S2V 1.0 uses the output of this tower directly instead of the Thinker hidden states at audio-token positions. A trainable audio encoder in the DiT adapts these features before the V2A conditioning path. The audio stream drives lip and head motion through the audio blocks and the V2A cross-attention layers.
  \item A single reference image $I_{\text{ref}}$ of the target identity. The image is mapped to the latent space by the Hunyuan VAE \cite{kong2025hunyuanvideosystematicframeworklarge} and patchified by the same visual tokenizer as the noisy video latents.
  \item An optional memory clip $V_{\text{mem}}$ of past frames, used to extend a generation beyond a single chunk of $125$ frames at $25$ FPS while helping preserve identity and motion continuity. The training schedule for the memory and the reference is detailed in Section~\ref{subsubsec:ref_memory_placement}.
\end{itemize}

During training, we sample one of six captions per clip: a short, medium, or long caption in English or Russian. The sampling probabilities are 0.075, 0.075, and 0.35 for the three respective granularity levels in each language. The speech waveform, the text prompt, and the reference image are each dropped independently during training with probability $0.1$ to support classifier-free guidance \cite{ho2022classifierfree}. The memory-free steps of Section~\ref{subsubsec:ref_memory_placement} train the model to generate without visual history, as required for the first chunk of every rollout.

\subsection{Latent video tokenization}

The visual stream operates on $5$D latents $z \in \mathbb{R}^{B \times T \times H \times W \times C}$, with batch size $B$, $T$ latent frames, latent height $H$ and width $W$, and $C$ channels, produced by the Hunyuan VAE with temporal compression ratio $4$ and spatial downsampling factor $8$. The training resolutions are $512 \times 512$, $768 \times 512$, and $512 \times 768$, with $125$ raw frames per clip producing $T = 32$ latent frames. The latent carries $16$ channels. The reference image contributes a single latent frame and the memory clip contributes up to $30$ latent frames. A target chunk of $32$ latent frames corresponds to $(32 - 1) \times 4 + 1 = 125$ video frames, that is $5$ seconds at $25$ FPS, so the memory reaches $30$ latent frames back, about $4.8$ seconds of history, using substantially fewer tokens than $30$ memory frames tokenized at the finest scale.

\begin{figure}[!ht]
\centering
\resizebox{\textwidth}{!}{%
\begin{tikzpicture}[
    font=\footnotesize,
    every node/.style={align=center, inner sep=4pt},
    framepack/.style={rectangle, rounded corners=2pt, draw, fill=cyan!15,
                      minimum height=0.6cm, minimum width=1.8cm, font=\scriptsize},
    cnode/.style={circle, draw, fill=white, minimum size=0.6cm, inner sep=0, font=\bfseries\small},
    attnV/.style={rectangle, rounded corners=2pt, draw, fill=cyan!18,
                  minimum height=2.9cm, minimum width=0.65cm, inner sep=1pt},
    arr/.style={-{Latex[length=2mm, width=1.5mm]}, semithick},
    lbl/.style={font=\scriptsize, inner sep=1pt},
]

\newcommand{\framestack}[2]{%
  \foreach \k in {2,1,0} {
    \node[draw, inner sep=0pt] at ($(#1) + (0.14*\k, 0.14*\k)$)
      {\includegraphics[width=1.9cm,height=1.25cm]{#2}};
  }%
}
\newcommand{\tokgrid}[3]{%
  \draw[fill=#3] ($(#1) + (-0.6, -0.6)$) rectangle ($(#1) + (0.6, 0.6)$);
  \pgfmathsetmacro{\step}{1.2/#2}
  \draw[step=\step cm, black!45, very thin, shift={($(#1) + (-0.6, -0.6)$)}] (0,0) grid (1.2,1.2);
  \draw ($(#1) + (-0.6, -0.6)$) rectangle ($(#1) + (0.6, 0.6)$);%
}

\coordinate (b3) at (0, 3.2);
\coordinate (b2) at (3.0, 3.2);
\coordinate (b1) at (6.0, 3.2);
\framestack{b3}{frames/f0.jpg}
\framestack{b2}{frames/f1.jpg}
\framestack{b1}{frames/f2.jpg}
\node[lbl, anchor=south] at ($(b3) + (0.14, 1.1)$) {Bucket 3\\$24$ latent frames};
\node[lbl, anchor=south] at ($(b2) + (0.14, 1.1)$) {Bucket 2\\$4$ latent frames};
\node[lbl, anchor=south] at ($(b1) + (0.14, 1.1)$) {Bucket 1\\$2$ latent frames};

\coordinate (g3) at (0, 0.7);
\coordinate (g2) at (3.0, 0.7);
\coordinate (g1) at (6.0, 0.7);
\tokgrid{g3}{2}{cyan!15}
\tokgrid{g2}{4}{cyan!15}
\tokgrid{g1}{8}{cyan!15}
\draw[arr] ($(b3) + (0, -0.75)$) -- node[lbl, right, xshift=2pt] {patch $(4,8,8)$} ($(g3) + (0, 0.65)$);
\draw[arr] ($(b2) + (0, -0.75)$) -- node[lbl, right, xshift=2pt] {patch $(2,4,4)$} ($(g2) + (0, 0.65)$);
\draw[arr] ($(b1) + (0, -0.75)$) -- node[lbl, right, xshift=2pt] {patch $(1,2,2)$} ($(g1) + (0, 0.65)$);
\node[lbl, anchor=north] at ($(g3) + (0, -0.7)$) {$6$ token frames};
\node[lbl, anchor=north] at ($(g2) + (0, -0.7)$) {$2$ token frames};
\node[lbl, anchor=north] at ($(g1) + (0, -0.7)$) {$2$ token frames};

\node[cnode] (c32) at (1.5, 0.7) {C};
\node[cnode] (c21) at (4.5, 0.7) {C};

\node[draw=Green, very thick, dashed, rounded corners=3pt,
      fit={($(b3) + (-1.2, 1.95)$) ($(g1) + (1.95, -1.15)$)}, inner sep=2pt,
      label={[font=\scriptsize\bfseries, text=Green, anchor=south]north:Frame pack memory $V_{\text{mem}}$}] (fpbox) {};

\node[draw, inner sep=0pt] (ref) at (9.8, 3.25) {\includegraphics[width=1.9cm,height=1.25cm]{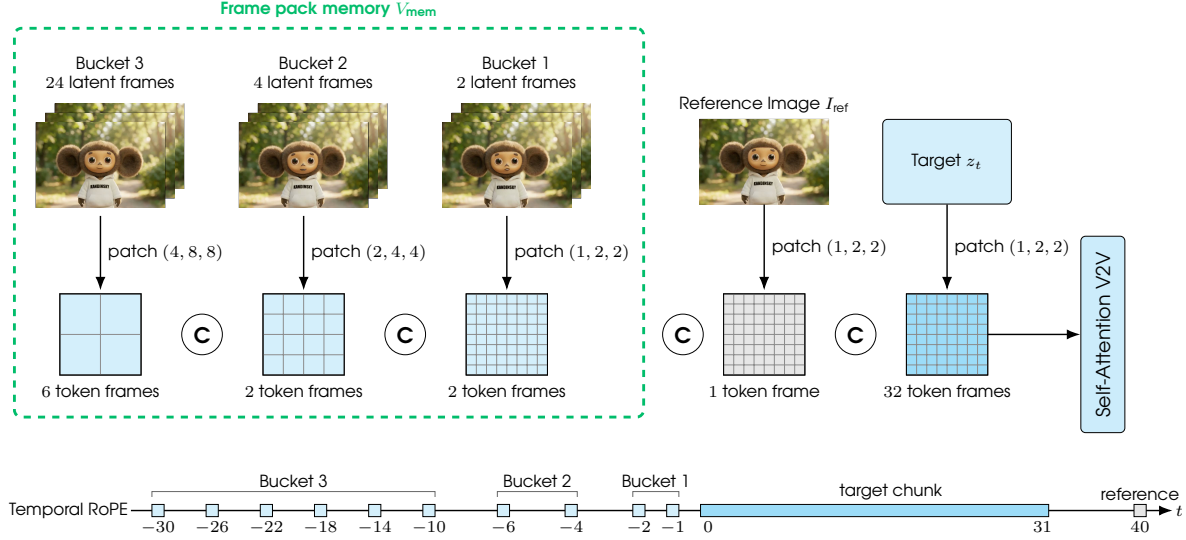}};
\node[lbl, anchor=south] at ($(ref.north) + (0, 0.05)$) {Reference Image $I_{\text{ref}}$};
\coordinate (gr) at (9.8, 0.7);
\tokgrid{gr}{8}{gray!20}
\draw[arr] (ref.south) -- node[lbl, right, xshift=2pt] {patch $(1,2,2)$} ($(gr) + (0, 0.65)$);
\node[lbl, anchor=north] at ($(gr) + (0, -0.7)$) {$1$ token frame};

\node[framepack, minimum width=1.9cm, minimum height=1.25cm] (tgt) at (12.5, 3.25) {Target $z_t$};
\coordinate (gt) at (12.5, 0.7);
\tokgrid{gt}{8}{cyan!35}
\draw[arr] (tgt.south) -- node[lbl, right, xshift=2pt] {patch $(1,2,2)$} ($(gt) + (0, 0.65)$);
\node[lbl, anchor=north] at ($(gt) + (0, -0.7)$) {$32$ token frames};

\node[cnode] (cm) at (8.6, 0.7) {C};
\node[cnode] (crt) at (11.15, 0.7) {C};

\node[attnV] (sa) at (14.8, 0.7) {\rotatebox{90}{Self-Attention V2V}};
\draw[arr] ($(gt) + (0.6, 0)$) -- (sa.west);

\def\ax{-1.9}
\newcommand{\tx}[1]{\fpeval{(#1 <= -7) ? (0.85 + (#1 + 30) * 0.2) : ((#1 <= 0) ? (5.45 + (#1 + 7) * 0.5) : (8.95 + (#1) * 0.16))}}
\draw[-{Latex[length=2mm, width=1.5mm]}, semithick] (\tx{-32}, \ax) -- (\tx{43}, \ax)
  node[lbl, right] {$t$};
\foreach \t in {-30,-26,-22,-18,-14,-10,-6,-4,-2,-1}
  \draw[fill=cyan!15] (\tx{\t} - 0.09, \ax - 0.09) rectangle (\tx{\t} + 0.09, \ax + 0.09);
\draw[fill=cyan!35] (\tx{0} - 0.09, \ax - 0.09) rectangle (\tx{31} + 0.09, \ax + 0.09);
\draw[fill=gray!20] (\tx{40} - 0.09, \ax - 0.09) rectangle (\tx{40} + 0.09, \ax + 0.09);
\foreach \t in {-30,-26,-22,-18,-14,-10,-6,-4,-2,-1,31,40}
  \node[lbl, anchor=north] at (\tx{\t}, \ax - 0.12) {$\t$};
\node[lbl, anchor=north west] at (\tx{0} - 0.09, \ax - 0.12) {$0$};
\draw[black!60] (\tx{-30} - 0.09, \ax + 0.2) -- ++(0, 0.08) -| (\tx{-10} + 0.09, \ax + 0.2);
\draw[black!60] (\tx{-6} - 0.09, \ax + 0.2) -- ++(0, 0.08) -| (\tx{-4} + 0.09, \ax + 0.2);
\draw[black!60] (\tx{-2} - 0.09, \ax + 0.2) -- ++(0, 0.08) -| (\tx{-1} + 0.09, \ax + 0.2);
\node[lbl, anchor=south] at (\tx{-20}, \ax + 0.3) {Bucket 3};
\node[lbl, anchor=south] at (\tx{-5}, \ax + 0.3) {Bucket 2};
\node[lbl, anchor=south] at (\tx{-1.5}, \ax + 0.3) {Bucket 1};
\node[lbl, anchor=south] at (\tx{17}, \ax + 0.12) {target chunk};
\node[lbl, anchor=south] at (\tx{40}, \ax + 0.12) {reference};
\node[lbl, anchor=east] at (\tx{-32}, \ax) {Temporal RoPE};

\end{tikzpicture}%
}
\caption{FramePack memory layout. The $30$ latent frames of the memory clip $V_{\text{mem}}$ are split into three buckets of $24$, $4$, and $2$ frames, from oldest to most recent, and patchified with patch sizes $(4, 8, 8)$, $(2, 4, 4)$, and $(1, 2, 2)$, so older frames contribute fewer tokens. The memory tokens are concatenated (C) with the reference and target tokens into one visual sequence for the DiT. The bottom axis shows the temporal RoPE position of each token group and is not drawn to scale.}
\label{fig:framepack}
\end{figure}

Latents are split into non-overlapping spatiotemporal patches of size $(1, 2, 2)$ before being projected to the model dimension. This patch size is used for the main noisy latents, the reference latent, and the $2$ most recent memory latents. The preceding $4$ memory latents use patches of size $(2, 4, 4)$. The oldest $24$ memory latents use patches of size $(4, 8, 8)$. Each memory scale uses its own linear projection, with weights separate from those of the main visual projection. Spatial RoPE positions of the two coarse scales are subsampled by the same factor as their patch size, so one coarse token covers a larger region of the frame. This three-scale layout adapts FramePack \cite{zhang2025frame} to compressed long-context video memory. Across all scales the rotary frequencies on the two spatial axes are divided by $2.0$, while the temporal axis is not rescaled. Figure~\ref{fig:framepack} shows the resulting memory layout and the temporal RoPE positions of all visual token groups.

\subsection{DiT backbone}

The backbone is a 3D Diffusion Transformer \cite{peebles2023dit} in which a text encoder and an audio encoder prepare the prompt and speech features, and a visual decoder generates the video and attends to both through cross-attention.

\paragraph{Text encoder.} A trainable stack of $N_t$ text transformer blocks with self-attention, RoPE \cite{su2023roformerenhancedtransformerrotary} positional encoding in $1$D, RMSNorm on queries and keys, and AdaLN-zero modulation \cite{peebles2023dit} driven by the diffusion time embedding combined with the pooled CLIP sentence embedding. We use $N_t = 2$ for the 2B model and $N_t = 4$ for the 21B model.

\paragraph{Audio encoder.} A trainable stack of $N_a$ unconditional transformer blocks with self-attention and $1$D RoPE on the audio token sequence. We use $N_a = 2$ for the 2B model and $N_a = 4$ for the 21B model. Audio blocks are not modulated by the diffusion time, because the speech track is the same across all denoising steps.

\paragraph{Visual decoder.} A stack of $N_v$ visual transformer blocks. Each block executes, in order, self-attention with $3$D RoPE on visual tokens, cross-attention to text, optional cross-attention to audio, and a feed-forward layer. The self-attention, text cross-attention, and FFN are gated by AdaLN-zero parameters predicted from the diffusion time embedding combined with the pooled CLIP sentence embedding, while the audio cross-attention is added as a residual without modulation. The 2B model uses $N_v = 32$ blocks with $d_{\text{model}} = 1792$, $d_{\text{ff}} = 7168$, and per-head RoPE dimension allocations of $(16, 24, 24)$, summing to a head dimension of $64$. The 21B model uses $N_v = 60$ blocks with $d_{\text{model}} = 4096$, $d_{\text{ff}} = 16384$, and per-head RoPE dimension allocations of $(32, 48, 48)$, summing to a head dimension of $128$. The time-embedding dimension is $512$ for the 2B model and $1024$ for the 21B model. Table~\ref{tab:variant_configs} collects the settings that differ between the two variants.

\begin{table}[H]
\centering
\caption{Settings that differ between the two variants. Both share the patch size $(1, 2, 2)$, the $16$-channel VAE latent, the $3584$-dimensional text and audio features from Qwen2.5-Omni, the $768$-dimensional pooled CLIP vector, the reference RoPE position $40$, and the memory frame buckets $(2, 4, 24)$.}
\label{tab:variant_configs}
\footnotesize
\begin{tabular}{lcc}
\toprule
Setting & 2B Lite & 21B Pro \\
\midrule
Parameters & $2.3$B & $21.4$B \\
Visual blocks $N_v$ & $32$ & $60$ \\
Text blocks $N_t$ & $2$ & $4$ \\
Audio blocks $N_a$ & $2$ & $4$ \\
Model dimension $d_{\text{model}}$ & $1792$ & $4096$ \\
Feed-forward dimension $d_{\text{ff}}$ & $7168$ & $16384$ \\
Time-embedding dimension & $512$ & $1024$ \\
Per-head RoPE axes $(t, h, w)$ & $(16, 24, 24)$ & $(32, 48, 48)$ \\
Head dimension & $64$ & $128$ \\
Blocks with audio cross-attention & $13$ & $18$ \\
\bottomrule
\end{tabular}
\end{table}

The memory, reference, and target tokens are joined into one sequence, and a learned embedding added to each token tells the model which of the three groups it belongs to. AdaLN-zero modulation for the reference and memory groups is computed from a time embedding evaluated at $t = 0$ rather than at the sampled diffusion time, so those tokens are always modulated as clean latents while the target tokens follow the noise level of the step. The temporal RoPE positions assigned to each sub-stream and the training schedule that mixes the memory and reference are described next.

\subsection{Reference image and memory placement}\label{subsubsec:ref_memory_placement}

We use two training regimes that differ in memory availability $V_{\text{mem}}$ and reference-frame selection $I_{\text{ref}}$ from the source video. With probability $0.75$ a step is sampled with the memory latents zeroed out while their tokens stay in the sequence, matching the memory configuration used for the first chunk at inference, and the reference frame is taken from the source video immediately before the chunk that the model is asked to denoise. With probability $0.25$ the step uses a memory clip, and the reference frame is taken from a position before the memory, far in the past relative to the denoised chunk. This second regime trains the model to use preceding frames as visual memory. At inference, the memory consists of previously generated frames.

On the temporal RoPE axis the current denoised chunk occupies positions $0$ to $T - 1$ with $T = 32$ latent frames. The reference image is fixed at the RoPE position $40$, well outside the chunk range. The memory clip, when present, is placed at negative RoPE positions before the chunk: $-2$ and $-1$ for the two fine latent frames, $-6$ and $-4$ for the medium scale, and $-30$ to $-10$ in steps of $4$ for the coarse scale. We use the mixed training schedule and explicit RoPE separation to discourage the model from reusing the reference pose at chunk boundaries. Such behavior can arise when the reference is always placed inside the target chunk during training.

\subsection{Audio injection}

\begin{figure}[!ht]
\centering
\resizebox{0.85\textwidth}{!}{%
\begin{tikzpicture}[
    font=\footnotesize,
    every node/.style={align=center, inner sep=4pt},
    inp/.style={rectangle, rounded corners=2pt, draw, fill=gray!10,
                minimum height=0.7cm, minimum width=1.9cm},
    frozen/.style={rectangle, rounded corners=2pt, draw, dashed, fill=gray!18,
                   minimum width=0.95cm, font=\scriptsize, inner sep=2pt},
    audenc/.style={rectangle, rounded corners=2pt, draw, fill=orange!20,
                   minimum height=0.85cm, minimum width=2.5cm},
    atok/.style={rectangle, rounded corners=1pt, draw, fill=orange!25,
                 minimum size=0.32cm, inner sep=0pt},
    vtok/.style={rectangle, rounded corners=1pt, draw, fill=cyan!25,
                 minimum size=0.32cm, inner sep=0pt},
    cellon/.style={rectangle, draw, fill=orange!55, minimum size=0.95cm, inner sep=0pt},
    celloff/.style={rectangle, draw=black!40, fill=white, minimum size=0.95cm, inner sep=0pt},
    note/.style={rectangle, rounded corners=2pt, draw, dashed, fill=gray!10,
                 inner sep=4pt, font=\scriptsize, align=center},
    arr/.style={-{Latex[length=2mm, width=1.5mm]}, semithick},
    lbl/.style={font=\scriptsize, inner sep=1pt},
]

\node[inp, minimum height=1.0cm, minimum width=2.0cm, inner sep=0pt] (aud) at (-3.3, 3.7) {};
\foreach \i/\h in {0/0.12, 1/0.28, 2/0.14, 3/0.36, 4/0.20, 5/0.28, 6/0.14, 7/0.32, 8/0.18, 9/0.12}
  \draw[line width=1pt, Green] (-3.75 + \i*0.10, 3.95 - \h/2) -- (-3.75 + \i*0.10, 3.95 + \h/2);
\node[font=\footnotesize] at (-3.3, 3.45) {Audio $a$};
\node[frozen, minimum height=1.0cm, minimum width=2.4cm] (qa) at (-0.35, 3.7) {Qwen2.5-Omni\\audio tower};
\node[audenc] (audE) at (2.65, 3.7) {Audio Encoder\\\scriptsize $N_a$ blocks};
\draw[arr] (aud.east) -- (qa.west);
\draw[arr] (qa.east) -- (audE.west);

\def\gx{9.0}   
\def\gy{2.6}   
\def\gs{1.25}  

\foreach \j/\lab in {1/$0$, 2/$1$, 3/$\cdots$, 4/$T-1$} {
  \pgfmathsetmacro{\cx}{\gx + (\j - 1) * \gs}
  \node[atok] at (\cx - 0.2, \gy + 1.1) {};
  \node[atok] at (\cx + 0.2, \gy + 1.1) {};
  \draw[black!60] (\cx - 0.42, \gy + 1.38) -- (\cx - 0.42, \gy + 1.46) -- (\cx + 0.42, \gy + 1.46) -- (\cx + 0.42, \gy + 1.38);
  \node[lbl, anchor=south] at (\cx, \gy + 1.48) {\lab};
}
\node[lbl, font=\scriptsize\bfseries, text=orange!80!black, anchor=south] at (\gx + 1.5 * \gs, \gy + 1.85) {Audio tokens (keys, values)};
\draw[arr] (audE.east) -- ($(\gx - 0.42, \gy + 1.1)$);

\foreach \i/\lab/\img in {1/$0$/f0, 2/$1$/f1, 3/$\vdots$/f2, 4/$T-1$/f3} {
  \pgfmathsetmacro{\cy}{\gy - (\i - 1) * \gs}
  \node[vtok] at (\gx - 1.1, \cy + 0.2) {};
  \node[vtok] at (\gx - 1.1, \cy - 0.2) {};
  \draw[black!60] (\gx - 1.38, \cy - 0.42) -- (\gx - 1.46, \cy - 0.42) -- (\gx - 1.46, \cy + 0.42) -- (\gx - 1.38, \cy + 0.42);
  \node[lbl, anchor=east] at (\gx - 1.48, \cy) {\lab};
  \node[draw, inner sep=0pt, anchor=east] at (\gx - 2.3, \cy) {\includegraphics[width=1.2cm,height=0.8cm]{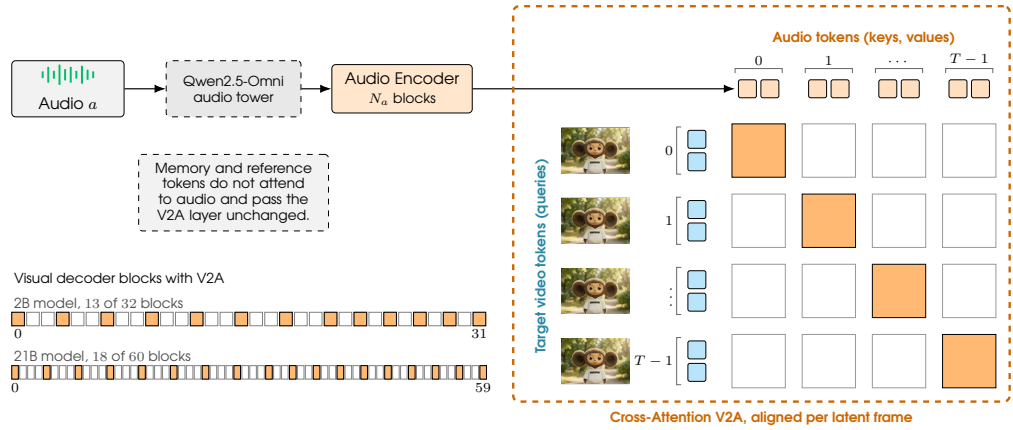}};
}
\node[lbl, font=\scriptsize\bfseries, text=cyan!60!black, rotate=90, anchor=south] at (\gx - 3.7, \gy - 1.5 * \gs) {Target video tokens (queries)};

\foreach \i in {1,...,4} {
  \foreach \j in {1,...,4} {
    \pgfmathsetmacro{\cx}{\gx + (\j - 1) * \gs}
    \pgfmathsetmacro{\cy}{\gy - (\i - 1) * \gs}
    \ifnum\i=\j
      \node[cellon] at (\cx, \cy) {};
    \else
      \node[celloff] at (\cx, \cy) {};
    \fi
  }
}

\node[draw=orange!80!black, very thick, dashed, rounded corners=3pt,
      fit={($(\gx - 4.3, \gy + 2.5)$) ($(\gx + 3 * \gs + 0.65, \gy - 3 * \gs - 0.65)$)}, inner sep=2pt,
      label={[font=\scriptsize\bfseries, text=orange!80!black, anchor=north]south:Cross-Attention V2A, aligned per latent frame}] (v2abox) {};

\node[note, text width=3.1cm] (bypass) at (-0.35, 1.85)
  {Memory and reference tokens do not attend to audio and pass the V2A layer unchanged.};

\node[lbl, anchor=south west] at (-4.3, 0.2) {Visual decoder blocks with V2A};
\def\sa{-0.4}
\def\pa{0.265}
\node[lbl, anchor=south west, text=black!70] at (-4.3, \sa + 0.14) {2B model, $13$ of $32$ blocks};
\foreach \k in {0,...,31}
  \draw[black!50, fill=white] (-4.3 + \k*\pa, \sa - 0.12) rectangle (-4.3 + \k*\pa + 0.225, \sa + 0.12);
\foreach \k in {0,3,6,9,12,15,18,21,23,25,27,29,31}
  \draw[fill=orange!55] (-4.3 + \k*\pa, \sa - 0.12) rectangle (-4.3 + \k*\pa + 0.225, \sa + 0.12);
\node[lbl, anchor=north] at (-4.3 + 0.1125, \sa - 0.14) {$0$};
\node[lbl, anchor=north] at (-4.3 + 31*\pa + 0.1125, \sa - 0.14) {$31$};
\def\sb{-1.35}
\def\pb{0.1413}
\node[lbl, anchor=south west, text=black!70] at (-4.3, \sb + 0.14) {21B model, $18$ of $60$ blocks};
\foreach \k in {0,...,59}
  \draw[black!50, fill=white] (-4.3 + \k*\pb, \sb - 0.12) rectangle (-4.3 + \k*\pb + 0.117, \sb + 0.12);
\foreach \k in {0,4,8,12,16,20,24,28,32,35,38,41,44,47,50,53,56,59}
  \draw[fill=orange!55] (-4.3 + \k*\pb, \sb - 0.12) rectangle (-4.3 + \k*\pb + 0.117, \sb + 0.12);
\node[lbl, anchor=north] at (-4.3 + 0.0585, \sb - 0.14) {$0$};
\node[lbl, anchor=north] at (-4.3 + 59*\pb + 0.0585, \sb - 0.14) {$59$};

\end{tikzpicture}%
}
\caption{Frame-aligned audio cross-attention (V2A). Audio features from the frozen Qwen2.5-Omni audio tower pass through the trainable audio encoder. In each V2A layer the target video tokens of latent frame $i$ attend only to the audio tokens of the same time span, which gives a block-diagonal attention pattern. Memory and reference tokens skip this layer. The strips at the bottom left mark the visual decoder blocks that contain a V2A layer in the 2B and 21B models.}
\label{fig:v2a}
\end{figure}

Audio is injected into the visual stream only at a subset of decoder blocks spread over the depth of the network. The 2B model injects audio at $13$ of $32$ blocks, at indices $0, 3, 6, 9, 12, 15, 18, 21, 23, 25, 27, 29, 31$ counting from zero. The 21B model injects audio at $18$ of $60$ blocks, at indices $0, 4, 8, 12, 16, 20, 24, 28, 32, 35, 38, 41, 44, 47, 50, 53, 56, 59$. The spacing narrows toward the output, from three blocks to two in the 2B model and from four to three in the 21B model, so a larger share of audio-conditioned blocks sits close to the output projection. At each audio-enabled block, only the current visual latents attend to audio, while reference and memory tokens are excluded from V2A cross-attention and pass through it unchanged. The V2A output is added to the visual latents as a residual, without AdaLN modulation. The V2A cross-attention is aligned per latent frame: visual tokens belonging to the same latent frame attend to the corresponding slice of the audio token sequence. Figure~\ref{fig:v2a} illustrates this attention pattern.

\section{Training Stages}\label{sec:training_stages}

\subsection{Continuous Pre-training}\label{subsec:model_training}

\subsubsection{Flow-matching objective}\label{subsubsec:flow_matching}

The model is trained with rectified-flow matching \cite{lipman2023flow, liu2022flowstraightfastlearning, esser2024scalingrectifiedflowtransformers}. Let $z_0$ be the clean video latent and $\epsilon \sim \mathcal{N}(0, I)$. The noisy latent at time $t \in [0, 1]$ is
$$z_t = (1 - t)\, z_0 + t\, \epsilon,$$
and the target velocity is $v(z_t, t) = \epsilon - z_0$. The network $f_\theta$ predicts $\hat{v} = f_\theta(z_t, t, c)$, where $c$ collects the text, audio, reference, and memory conditioning. The training loss is the mean squared error
$$\mathcal{L}_{\text{flow}} = \mathbb{E}_{t, \epsilon, c, z_0} \left[ \| f_\theta(z_t, t, c) - v(z_t, t) \|_2^2 \right].$$

Timesteps are sampled from a reweighted logit-normal distribution. We first draw $u = \sigma(\eta)$ with $\eta \sim \mathcal{N}(0, 1)$ and the logistic sigmoid $\sigma(x) = 1 / (1 + e^{-x})$, then push mass toward high noise by
$$t = \frac{s\, u}{1 + (s - 1)\, u}.$$
This is the scaled logit-normal schedule of \cite{esser2024scalingrectifiedflowtransformers} with scale $s = 5$. The transformation allocates more training samples to high-noise states than the unscaled logit-normal distribution.

\subsubsection{Face loss}\label{subsubsec:face_loss}

In addition to the flow-matching MSE we use a region-weighted auxiliary loss on the face. For each clip, a face mask $M$ aligned to the latent grid identifies the speaker’s face. The face loss is
$$\mathcal{L}_{\text{face}} = \frac{\sum (f_\theta(z_t, t, c) - v)^2 \odot M}{\sum M + \delta},$$
where $\delta = 10^{-8}$ prevents division by zero in the masked mean, and the total loss is $\mathcal{L} = \mathcal{L}_{\text{flow}} + \lambda_{\text{face}} \mathcal{L}_{\text{face}}$. For $\lambda_{\text{face}}$ we use an adaptive weighting based on the inverse of the face-area ratio of the clip,
$$\lambda_{\text{face}} = \operatorname{clip}\!\left(\frac{r_{\text{ref}}}{r_{\text{face}}},\; \lambda_{\min},\; \lambda_{\max}\right), \qquad r_{\text{face}} = \frac{\sum M}{|M|},$$
where $r_{\text{ref}}$ is the reference face-area ratio, $r_{\text{face}}$ is the face-area ratio of the clip (the fraction of the mask occupied by the face), and $\lambda_{\min}, \lambda_{\max}$ bound the resulting weight. The constants $r_{\text{ref}} = 0.04$, $\lambda_{\min} = 0.15$, and $\lambda_{\max} = 2.5$ are chosen in accordance with the empirical face-area distribution of the training data. We use the masked mean form, which keeps the gradient on the face region stable when the face occupies only a small fraction of the frame. With a fixed $\lambda_{\text{face}}$, clips with large faces can dominate the face gradient. In such runs we observed that the model tended to frame the face more tightly as training progressed. The adaptive weight is meant to balance the face term across face sizes and keep a useful signal when the face is small in the frame.

\subsubsection{Pretraining initialization}

The DiT for Kandinsky S2V 1.0 is initialized from the corresponding Kandinsky 5.0 text-to-video checkpoints \cite{kandinskylab2025kandinsky5}, which were pretrained at $512$ and $768$ resolution without audio conditioning. The text and audio encoders in the DiT, the audio cross-attention layers, the memory projections, and the reference-image segment embedding are initialized from scratch. The Qwen2.5-Omni Thinker and audio tower, the CLIP text encoder, and the VAE remain frozen throughout training. All DiT blocks and the newly introduced conditioning layers are trainable.

\subsubsection{Multi-resolution training}

Each step samples a single resolution from $\{(512, 512), (768, 512), (512, 768)\}$, while the frame count stays fixed at $125$. Both model sizes use dense FlashAttention \cite{dao2022flash, dao2023flash2} throughout. A resolution-aware batch accumulator keeps microbatch shapes consistent within each gradient accumulation step, as required by the compiled FSDP graphs in our implementation.

\subsubsection{Optimization}

Both model sizes are trained with AdamW at learning rate $3 \times 10^{-5}$, weight decay $10^{-3}$, $\beta = (0.9, 0.95)$, $\epsilon_{\text{Adam}} = 10^{-8}$, and global gradient-norm clipping at $1.0$. The learning rate is constant after linear warmup over $200$ steps for the 2B model and $500$ steps for the 21B model. An exponential moving average of the weights with decay $0.999$ is maintained for inference. Mixed precision uses bfloat16 parameters and outputs with float32 gradient reduction. AdaLN-zero modulations and time embeddings are computed in float32 for numerical stability.

\subsubsection{Distributed training}

The 2B model is trained on $64$ GPUs, with FSDP \cite{zhao2023pytorchfsdpexperiencesscaling} sharding the weights $8$-way inside a node and replicating $8$-way across nodes. The 21B model is trained on $32$ nodes of $8$ GPUs, for a total of $256$ devices, with a $64$-way FSDP mesh and a $2$-way tensor-parallel mesh on the DiT, and a $64$-way FSDP mesh on the embedders. Tensor parallelism shards the visual sequence along height, so each rank holds a spatial slice of the latent and the ranks synchronize inside attention. Activation checkpointing \cite{activation_checkpointing} is applied to every visual decoder block. The text and audio encoders run only on the first rank of each tensor-parallel group, while the remaining tensor-parallel ranks receive the resulting embeddings via collective broadcast, which avoids duplicating the frozen Qwen2.5-Omni forward pass.

\subsubsection{Training setup}

Table~\ref{tab:training_setup} collects the settings of the two runs. The 2B model runs $14000$ steps at a global batch of $128$, the 21B Pro model $9500$ steps at a global batch of $128$. Both runs use the adaptive masked-mean face loss.

\begin{table}[H]
\centering
\caption{Training setup of the two runs. Both use AdamW with $\beta = (0.9, 0.95)$, $\epsilon_{\text{Adam}} = 10^{-8}$, gradient-norm clipping at $1.0$, an EMA decay of $0.999$, a timestep shift of $s = 5$, dense FlashAttention, independent text, audio, and reference dropout at probability $0.1$, and a memory-conditioned step probability of $0.25$.}
\label{tab:training_setup}
\footnotesize
\begin{tabular}{lcc}
\toprule
Setting & 2B & 21B Pro \\
\midrule
Initialization & \multicolumn{2}{c}{Kandinsky 5.0 Video T2V, EMA weights} \\
Training steps & $14000$ & $9500$ \\
Global batch size & $128$ & $128$ \\
Samples per data-parallel rank & $2$ & $1$ \\
Nodes $\times$ GPUs & $8 \times 8$ & $32 \times 8$ \\
FSDP shard width & $8$ & $64$ \\
Tensor-parallel width & $1$ & $2$ \\
\midrule
Learning rate & $3 \times 10^{-5}$ & $3 \times 10^{-5}$ \\
Weight decay & $10^{-3}$ & $10^{-3}$ \\
Warmup steps & $200$ & $500$ \\
Face loss & \multicolumn{2}{c}{Adaptive weight, masked mean} \\
\midrule
Target latent frames & $32$ & $32$ \\
Video frames per chunk & $125$ & $125$ \\
Resolutions & \multicolumn{2}{c}{$512 \times 512$, $768 \times 512$, $512 \times 768$} \\
\bottomrule
\end{tabular}
\end{table}

\subsection{Distillation}\label{sec:distillation}
To accelerate inference, we distill Kandinsky S2V 1.0 Lite and Pro using $\pi$-Flow \cite{chen2025piflow}, enabling generation in 16 sampling steps without classifier-free guidance. Both models follow the same distillation procedure.

\subsubsection{Method}\label{subsec:distill_method}

$\pi$-Flow replaces the velocity prediction of a few-step student with a policy: at the start of the $16$ segments of the noise schedule the network predicts, in a single forward pass, the parameters of a policy that is integrated across the segment using multiple substeps without further network calls. We use the dynamic-$\hat{x}_0$ (DX) policy of $\pi$-Flow~\cite{chen2025piflow}, in which the network predicts a grid of $5$ clean-latent estimates $\hat{z}_0$ at evenly spaced times within the segment, intermediate estimates are linearly interpolated, and the velocity follows from the flow-matching interpolation as $(z_t - \hat{z}_0) / t$ (Section~\ref{subsubsec:flow_matching}). The student is trained with policy-based imitation distillation ($\pi$-ID), the training algorithm of $\pi$-Flow. Each segment is split into three teacher steps of random length; at the start of each step the frozen teacher returns the guided velocity, and an $\ell_2$ loss pulls the average velocity of the policy over that step toward it, so that integrating the policy reproduces the teacher's displacement over the step. Architecturally the student is the teacher with the output projection widened five-fold. At initialization the teacher weights are replicated across the five grid points, so every grid estimate coincides and the policy reduces to the teacher velocity.

\paragraph{Guided teacher.} The imitation target is the guided velocity $v_u + w\,(v_c - v_u)$ with inference guidance scale $w = 3$ (Section~\ref{subsec:long_form_inference}). Here, $v_c$ and  $v_u$ denote the conditional and unconditional velocity predictions, respectively. The unconditional branch receives a fixed negative prompt and zeroed audio features, while the reference and memory latents are retained in both branches. The student therefore learns the guided flow directly and is sampled without guidance.

\paragraph{Inference schedule.} The policy grid is defined in raw segment time, while the network sees the transformed timestep (Section~\ref{subsubsec:flow_matching}). We use the same transform scale during distillation and inference, $s = 10$, rather than the $s = 5$ used during continuous pre-training, to align the distillation segments with the inference schedule.

\paragraph{Conditioning.} Text, audio, and reference dropout are switched off, so the student matches the teacher on the fully conditioned distribution it will see at inference. Memory dropout is retained, with its probability set to $0.5$ during distillation, so that the student retains the memory-free regime used for the first chunk. Distillation runs on single chunks with the memory clip taken from the data, exactly as in training, and the long-form chain is unchanged at inference (Section~\ref{subsec:long_form_inference}). The face loss (Section~\ref{subsubsec:face_loss}) is not used, since the target is the teacher velocity rather than the data.

\paragraph{Off-policy imitation.} $\pi$-ID is an on-policy method: the states at which the teacher is queried come from the student's own rollout. For teachers that are sensitive to off-trajectory states, $\pi$-Flow~\cite{chen2025piflow} adds scheduled trajectory mixing, which starts from states on the teacher trajectory, that is behavior cloning, and decays toward on-policy states. We stay at the first stage of that schedule for the whole run: states are advanced by the teacher's own guided velocity over each teacher step, and only the policy's average velocity over the step is learned.

\subsubsection{Training Setup}\label{subsec:distill_setup}

Table~\ref{tab:distill_settings} lists the settings of both runs. Each optimizer step requires one forward and one backward pass through the student and six forward passes through the frozen teacher: three teacher steps with two guidance branches each. A step takes about $140$ seconds for Pro on $64$ GPUs and about $60$ seconds for Lite on $32$ GPUs.

\begin{table}[H]
\centering
\caption{Settings of the $\pi$-Flow distillation of Kandinsky S2V 1.0 Pro and Lite. The upper block is shared by both runs; the lower block lists the settings that differ.}
\label{tab:distill_settings}
\footnotesize
\setlength{\tabcolsep}{6pt}
\begin{tabular}{lll}
\toprule
Setting & \multicolumn{2}{l}{Shared} \\
\midrule
Student NFE & \multicolumn{2}{l}{16} \\
Policy & \multicolumn{2}{l}{DX, 5 grid points} \\
Integration substeps (total) & \multicolumn{2}{l}{128} \\
Teacher steps per segment & \multicolumn{2}{l}{3} \\
Velocity-averaging window & \multicolumn{2}{l}{3 substeps} \\
Timestep shift & \multicolumn{2}{l}{10} \\
Teacher CFG scale & \multicolumn{2}{l}{3} \\
Text / audio / reference dropout & \multicolumn{2}{l}{0} \\
Memory dropout & \multicolumn{2}{l}{0.5} \\
Face loss & \multicolumn{2}{l}{Disabled} \\
Training clips & \multicolumn{2}{l}{9.5K} \\
Resolutions & \multicolumn{2}{l}{$512 \times 512$, $768 \times 512$, $512 \times 768$} \\
AdamW betas & \multicolumn{2}{l}{$(0.9, 0.95)$} \\
Weight decay & \multicolumn{2}{l}{0} \\
Warmup steps & \multicolumn{2}{l}{100} \\
EMA decay & \multicolumn{2}{l}{0.999} \\
Parameters / outputs & \multicolumn{2}{l}{BF16} \\
Gradient reduction & \multicolumn{2}{l}{FP32} \\
Global / per-rank batch size & \multicolumn{2}{l}{64 / 1} \\
Gradient accumulation steps & \multicolumn{2}{l}{2} \\
Optimizer steps & \multicolumn{2}{l}{$\sim$8,000} \\
Inference weights & \multicolumn{2}{l}{EMA} \\
\midrule
Setting & Pro & Lite \\
\midrule
Initialization & Pro EMA & Lite EMA \\
Learning rate & $2 \times 10^{-6}$ & $1.5 \times 10^{-6}$ \\
Gradient clipping & 5.0 & 10.0 \\
GPUs & 64 & 32 \\
Parallelism & FSDP + 2-way TP & 8-way FSDP \\
\bottomrule
\end{tabular}
\end{table}

In the Pro run, tensor parallelism shards the visual sequence along height, as in pre-training (Section~\ref{subsec:model_training}). Since the policy acts token-wise, this only requires the noisy states and segment boundaries to be shared within a tensor-parallel group.

\paragraph{Distillation data.} Distillation uses a curated set of $9.5$K single-speaker clips assembled for supervised fine-tuning, separate from the main corpus (Section~\ref{sec:data_processing_pipeline}), at the same three standard-definition resolutions as training. The videos were manually selected by annotators from open datasets and large video platforms based on the following criteria: overall visual quality, accurate lip synchronization, and diversity in speakers, spoken languages, emotions, and person placement within the frame. The selected videos then went through the same pipeline as the main corpus, active-speaker detection and clip extraction, quality filtering, and captioning in both languages (Section~\ref{sec:data_processing_pipeline}). Each training example consists of a $125$-frame target chunk, its associated memory clip, and six captions. The set passed the same exclusion of evaluation data as the main corpus, by source video identifier and by ArcFace identity (Section~\ref{subsec:provenance}).
\section{Inference}\label{subsec:long_form_inference}

\paragraph{Long-form generation.} A long video is a sequence of $125$-frame chunks. The first chunk is conditioned on  the reference image, the corresponding slice of the speech track, and the prompt, with the memory latents set to zero. Every later chunk reuses the same reference and fills the three memory scales from the generated history, taking the last $2$ latent frames at the target patch scale, the $4$ before them at $(2, 4, 4)$, and the $24$ before those at $(4, 8, 8)$. Memory is read from latents before decoding, avoiding VAE re-encoding of previously generated frames. The fixed history window keeps the visual-memory cache bounded as output length increases.

\paragraph{Sampling and guidance.} The base models use $50$ sampling steps at timestep shift $s = 10$. Each step requires a conditional and an unconditional model evaluation, for a total of $100$ NFE per chunk. Compared with the $s = 5$ used in training (Section~\ref{subsubsec:flow_matching}), $s = 10$ concentrates the steps in the high-noise region. The unconditional pass uses fixed negative prompt and zeroed audio features. Reference and memory latents stay in both branches, so identity conditioning is never guided. Writing $v_c$ and $v_u$ for the conditional and unconditional predicted velocities, the guided velocity is $v_u + w\,(v_c - v_u)$ with $w = 3$, so text and audio conditioning share one guidance scale.

\paragraph{Distilled sampling.} The distilled Lite and Pro models keep the chunking, reference, and memory handling above and use $\pi$-Flow sampling with $16$ policy segments and the same timestep shift $s = 10$ (Section~\ref{sec:distillation}). Each segment requires one forward pass, and the policy is integrated within it without further network calls. Guidance is absorbed during distillation, so there is no unconditional pass and no negative prompt, and a generation requires $16$ NFE instead of $100$.

\paragraph{Resolution and runtime.} Although Lite and Pro were trained only at SD resolutions, both can generate HD video at $25$ FPS without further training. Table~\ref{tab:chunk_sampling_time} compares sampling times measured after warm-up for one $125$-frame chunk, with \texttt{torch.compile} disabled. On eight GPUs, distillation reduces Pro's time from $162$ to $57$ seconds at SD and from $568$ to $145$ seconds at HD; on one GPU, Lite falls from $220$ to $43$ seconds at SD.

\begin{table}[H]
\centering
\caption{Per-chunk sampling times for Lite and Pro with the released inference code on H100 GPUs. GPU counts and times correspond in order within each row.}
\label{tab:chunk_sampling_time}
\small
\setlength{\tabcolsep}{7pt}
\begin{tabular}{llcc}
\toprule
Model & Resolution & GPUs & Sampling time per chunk (s) \\
\midrule
Kandinsky S2V 1.0 Lite, distilled & SD $512 \times 768$ & $1/2/4$ & $43/27/19$ \\
Kandinsky S2V 1.0 Lite & SD $512 \times 768$ & $1/2/4$ & $220/123/69$ \\
Kandinsky S2V 1.0 Pro, distilled & SD $512 \times 768$ & $1/2/4/8$ & $173/105/70/57$ \\
Kandinsky S2V 1.0 Pro & SD $512 \times 768$ & $1/2/4/8$ & $914/497/272/162$ \\
Kandinsky S2V 1.0 Pro, distilled & HD $1280 \times 768$ & $1/2/4/8$ & $675/375/219/145$ \\
Kandinsky S2V 1.0 Pro & HD $1280 \times 768$ & $1/2/4/8$ & $3854/2017/1060/568$ \\
\bottomrule
\end{tabular}
\end{table}

\paragraph{Distillation speedup.} In Table~\ref{tab:chunk_sampling_time} the speedup of distilled Pro at SD falls from $5.3\times$ on one GPU to $2.8\times$ on eight, because the released Pro configurations move the DiT weights onto every GPU before each chunk, and this transfer does not shrink with the number of GPUs. With the weights kept on the GPUs and \texttt{torch.compile} enabled, on one node of $8$ H100 80\,GB GPUs, a three-chunk SD video takes $554.9$ seconds with base Pro and $102.2$ seconds with distilled Pro, a $5.4\times$ end-to-end reduction, and a two-chunk HD video at $1280 \times 768$ takes $1563$ and $280$ seconds, or $5.6\times$. Both are below the $6.25\times$ ratio of NFE, $100$ against $16$, since VAE decoding, the frozen encoders, and video assembly are not accelerated.
\section[Evaluation]{Evaluation}\label{sec:evaluation}

\subsection{Comparison with Competing Models}\label{sec:auto_eval}

\begin{table}[!t]
\centering
\caption{Open-source comparison on EMTD, CelebV-HQ, and HDTF. The base and distilled Kandinsky models share one evaluation pool with the baselines. Mean rank and min-max aggregate the five reported metrics within each benchmark; best values are in bold.}
\label{tab:comparison}
\small
\setlength{\tabcolsep}{2.5pt}
\renewcommand{\arraystretch}{1.1}
\begin{tabular}{llrrrrrrr}
\toprule
Benchmark & Model & {\footnotesize Mean rank $\downarrow$} & Min-max $\uparrow$ & Sync-C $\uparrow$ & Sync-D $\downarrow$ & CSIM $\uparrow$ & IQA $\uparrow$ & ASE $\uparrow$ \\
\midrule
\multirow{11}{*}{EMTD} & Kandinsky S2V Pro (dist.) & {\bfseries 2.30} & {\bfseries 0.95} & 8.03 & 7.85 & {\bfseries 0.81} & 4.81 & 3.61 \\
 & Kandinsky S2V Pro & 3.90 & 0.82 & 8.05 & 7.62 & 0.77 & 4.73 & 3.54 \\
 & Kandinsky S2V Lite (dist.) & 4.90 & 0.87 & 7.41 & 8.24 & 0.79 & 4.81 & 3.58 \\
 & Kandinsky S2V Lite & 5.10 & 0.83 & 7.57 & 8.03 & 0.76 & 4.77 & 3.58 \\
 & LongCat 1.5 & 2.90 & 0.88 & {\bfseries 8.41} & {\bfseries 7.39} & 0.75 & 4.73 & {\bfseries 3.63} \\
 & Wan-S2V & 6.40 & 0.80 & 7.01 & 8.47 & 0.73 & {\bfseries 4.82} & 3.59 \\
 & HunyuanVideo-Avatar & 6.40 & 0.76 & 7.52 & 8.30 & 0.79 & 4.69 & 3.52 \\
 & OmniAvatar & 7.60 & 0.43 & 7.78 & 7.90 & 0.65 & 4.57 & 3.29 \\
 & InfiniteTalk & 7.70 & 0.58 & 7.33 & 7.93 & 0.74 & 4.58 & 3.41 \\
 & MultiTalk & 9.20 & 0.35 & 7.38 & 7.98 & 0.61 & 4.52 & 3.32 \\
 & StableAvatar & 9.60 & 0.19 & 3.12 & 12.10 & 0.62 & 4.69 & 3.41 \\
\midrule
\multirow{11}{*}{CelebV-HQ} & Kandinsky S2V Pro (dist.) & {\bfseries 2.40} & {\bfseries 0.93} & 7.19 & 8.15 & 0.78 & {\bfseries 4.23} & 3.15 \\
 & Kandinsky S2V Pro & 4.60 & 0.80 & 7.20 & 7.98 & 0.73 & 4.13 & 3.08 \\
 & Kandinsky S2V Lite (dist.) & 5.90 & 0.81 & 6.48 & 8.68 & 0.75 & 4.19 & 3.13 \\
 & Kandinsky S2V Lite & 5.10 & 0.80 & 6.78 & 8.30 & 0.75 & 4.13 & 3.10 \\
 & LongCat 1.5 & 2.60 & 0.89 & {\bfseries 7.55} & {\bfseries 7.73} & 0.74 & 4.16 & {\bfseries 3.16} \\
 & HunyuanVideo-Avatar & 5.20 & 0.66 & 6.98 & 8.14 & {\bfseries 0.80} & 3.87 & 2.87 \\
 & InfiniteTalk & 6.30 & 0.64 & 6.55 & 8.21 & 0.77 & 3.91 & 2.92 \\
 & Wan-S2V & 7.30 & 0.66 & 6.55 & 8.60 & 0.65 & 4.18 & 3.09 \\
 & OmniAvatar & 8.50 & 0.45 & 6.78 & 8.42 & 0.71 & 3.81 & 2.79 \\
 & StableAvatar & 8.90 & 0.24 & 2.99 & 11.78 & 0.73 & 3.94 & 2.88 \\
 & MultiTalk & 9.20 & 0.39 & 6.61 & 8.36 & 0.70 & 3.77 & 2.75 \\
\midrule
\multirow{11}{*}{HDTF} & Kandinsky S2V Pro (dist.) & {\bfseries 2.70} & {\bfseries 0.94} & 8.12 & 7.82 & 0.86 & {\bfseries 4.63} & {\bfseries 3.58} \\
 & Kandinsky S2V Pro & 5.80 & 0.81 & 8.01 & 7.78 & 0.84 & 4.51 & 3.51 \\
 & Kandinsky S2V Lite (dist.) & 5.00 & 0.89 & 7.60 & 8.18 & 0.86 & 4.61 & 3.57 \\
 & Kandinsky S2V Lite & 4.90 & 0.86 & 7.75 & 7.96 & 0.86 & 4.53 & 3.54 \\
 & LongCat 1.5 & 3.00 & 0.92 & {\bfseries 8.48} & {\bfseries 7.34} & 0.85 & 4.55 & 3.57 \\
 & HunyuanVideo-Avatar & 4.60 & 0.79 & 8.07 & 7.74 & {\bfseries 0.87} & 4.39 & 3.39 \\
 & InfiniteTalk & 6.40 & 0.63 & 7.70 & 7.59 & 0.86 & 4.21 & 3.24 \\
 & Wan-S2V & 6.80 & 0.79 & 7.49 & 8.10 & 0.83 & 4.60 & 3.52 \\
 & StableAvatar & 8.50 & 0.46 & 3.67 & 11.69 & 0.85 & 4.50 & 3.49 \\
 & MultiTalk & 8.60 & 0.35 & 7.74 & 7.77 & 0.79 & 4.10 & 3.10 \\
 & OmniAvatar & 9.70 & 0.49 & 7.25 & 8.51 & 0.82 & 4.25 & 3.24 \\
\bottomrule
\end{tabular}
\end{table}

Table~\ref{tab:comparison} reports a quantitative comparison on the semi-body EMTD benchmark \cite{meng2025echomimicv2} and the talking-head CelebV-HQ \cite{zhu2022celebvhq} and HDTF (10-second clips) \cite{zhang2021flow} benchmarks. We compare the base and distilled Kandinsky S2V 1.0 Lite (2B) and Kandinsky S2V 1.0 Pro (21B) models with LongCat 1.5 \cite{longcatteam2026longcatavatar15}, HunyuanVideo-Avatar \cite{chen2025hunyuanvideo}, MultiTalk \cite{kong2025multitalk}, InfiniteTalk \cite{yang2025infinitetalk}, OmniAvatar \cite{gan2025omniavatar}, Wan-S2V \cite{humanaigc2025wans2v}, and StableAvatar \cite{tu2025stableavatar}. Every system is run at standard definition. The human evaluations below include both Pro and Lite.

Sync-C and Sync-D are the confidence and embedding-distance outputs of SyncNet \cite{chung2016out}. Higher Sync-C and lower Sync-D indicate better agreement between speech and lip motion. CSIM is the cosine similarity between ArcFace embeddings \cite{deng2019arcface} of the reference image and generated frames, where higher values indicate better identity preservation. IQA and ASE are Q-Align \cite{wu2023qalign} scores for technical image quality and visual aesthetics. Higher values are better for both scores.

Each benchmark contains $50$ clips at $25$ FPS, prepared at $512 \times 512$, $768 \times 512$, or $512 \times 768$ according to the source aspect ratio. HDTF clips are selected for image quality and face size and cut to the first $10$ seconds. EMTD clips must be at least four seconds long, pass a quality threshold, and have sufficiently large faces; we use their first $30$ seconds. CelebV-HQ uses full-length talk-labeled clips that pass a ground-truth lip-sync screen and have no letterboxing. For each clip, we extract the same reference frame, driving audio, and ground-truth video for all models. Each model runs in its own environment, and metrics are computed separately, with videos resized when needed. Table entries report the mean score across clips.

The two aggregates in Table~\ref{tab:comparison} are computed separately within each benchmark over the eleven compared models. Mean rank averages a model's ranks across the five metrics, with rank $1$ assigned to the best value in each column; lower is better. Min-max rescales each metric column so that the best model scores $1$ and the worst $0$, then averages those five scores; higher is better. Both reverse the direction of Sync-D and use metric values before rounding, so cells printed with the same value can have different aggregate scores.

\noindent\begin{minipage}{\linewidth}
\setlength{\parindent}{1.5em}
Across EMTD, CelebV-HQ, and HDTF, distilled Kandinsky S2V 1.0 Pro ranks first by both mean rank and min-max score among the evaluated models, followed by LongCat 1.5. Distilled Lite ranks third by min-max score on all three benchmarks. Distilled Pro achieves the highest CSIM on EMTD, the highest IQA on CelebV-HQ and HDTF, and the highest ASE on HDTF. On Sync-C, Kandinsky S2V 1.0 Pro is outperformed only by LongCat 1.5 on each benchmark, in its base or distilled form.
\end{minipage}

\begin{figure}[H]
\centering
\begin{subfigure}{\textwidth}
\centering
\includegraphics[width=0.78\textwidth]{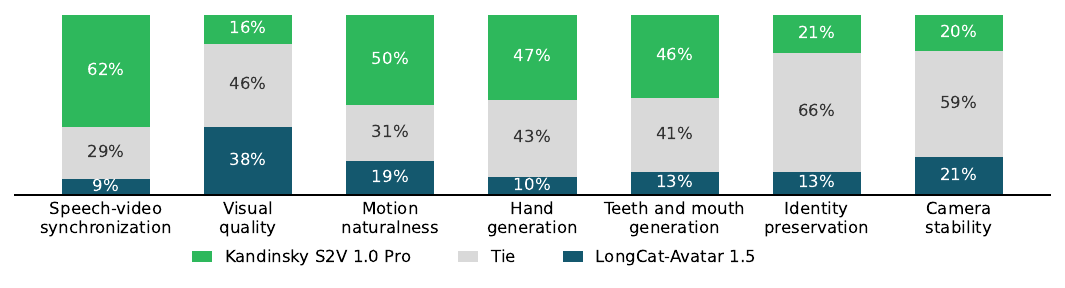}
\caption{Kandinsky S2V 1.0 Pro versus LongCat 1.5, SD.}
\end{subfigure}

\begin{subfigure}{\textwidth}
\centering
\includegraphics[width=0.78\textwidth]{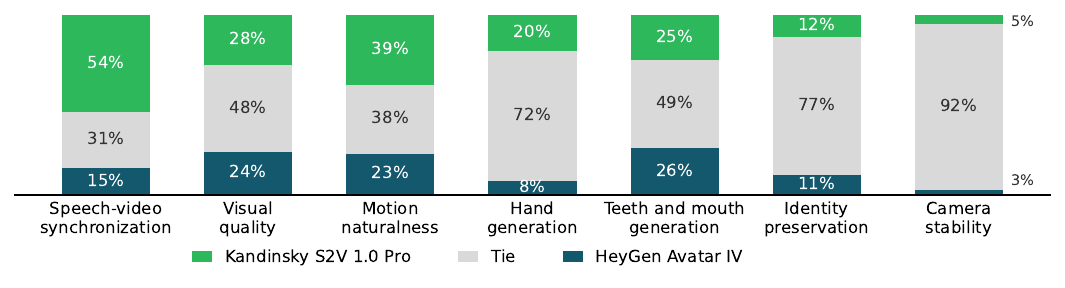}
\caption{Kandinsky S2V 1.0 Pro versus HeyGen Avatar IV, HD.}
\end{subfigure}

\begin{subfigure}{\textwidth}
\centering
\includegraphics[width=0.78\textwidth]{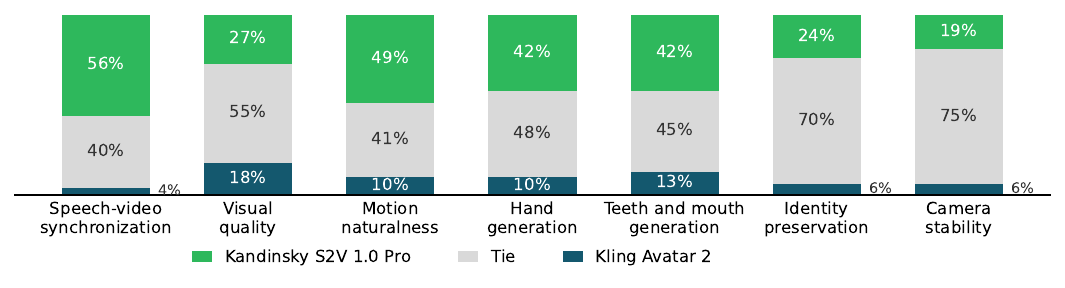}
\caption{Kandinsky S2V 1.0 Pro versus Kling Avatar 2, HD.}
\end{subfigure}
\caption{Side-by-side human evaluation on the $80$-item Russian set.}
\label{fig:sbs}
\end{figure}

\subsection{Human Evaluation}\label{sec:human_eval}

\begin{figure}[p]
\centering
\begin{subfigure}{\textwidth}
\centering
\includegraphics[width=0.78\textwidth]{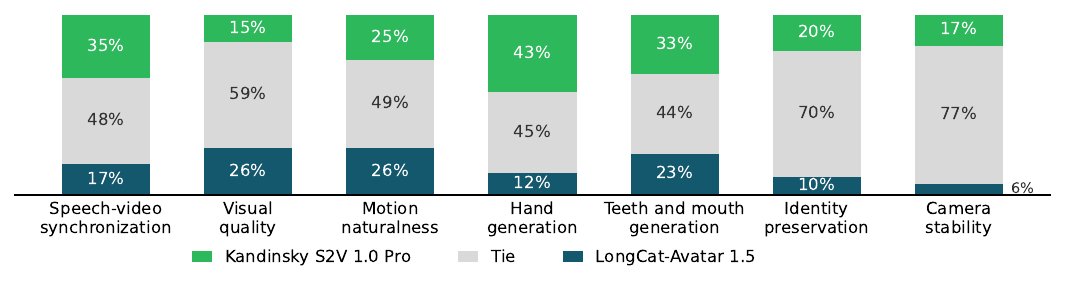}
\caption{Kandinsky S2V 1.0 Pro versus LongCat 1.5, HD.}
\end{subfigure}

\begin{subfigure}{\textwidth}
\centering
\includegraphics[width=0.78\textwidth]{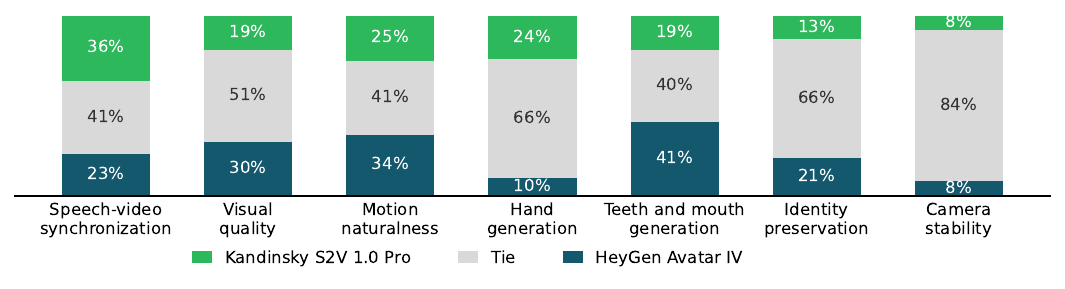}
\caption{Kandinsky S2V 1.0 Pro versus HeyGen Avatar IV, HD.}
\end{subfigure}
\caption{Side-by-side human evaluation on the multilingual set.}
\label{fig:sbs_intl}
\end{figure}

\begin{figure}[p]
\centering
\begin{subfigure}{\textwidth}
\centering
\includegraphics[width=0.78\textwidth, trim={0 -6.5bp 0 -6bp}]{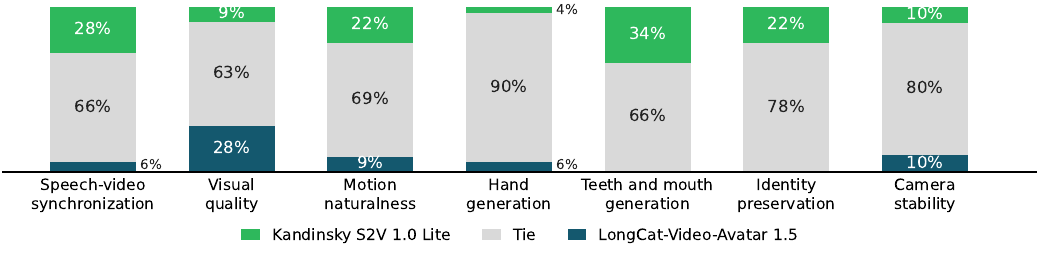}
\caption{Kandinsky S2V 1.0 Lite versus LongCat 1.5, SD.}
\end{subfigure}

\begin{subfigure}{\textwidth}
\centering
\includegraphics[width=0.78\textwidth, trim={0 -6.5bp 0 -6bp}]{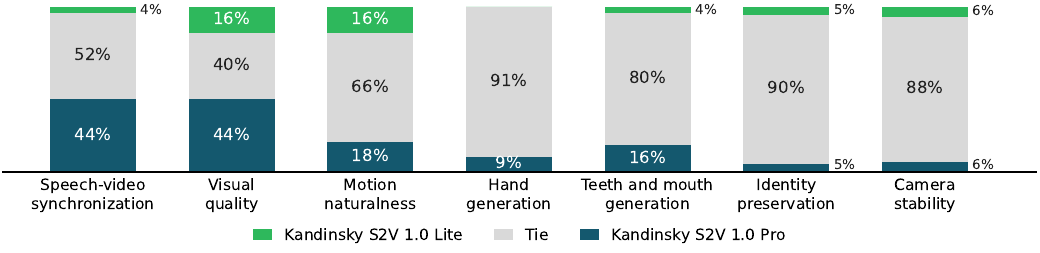}
\caption{Kandinsky S2V 1.0 Lite versus Pro, SD.}
\end{subfigure}
\caption{Additional side-by-side human evaluations involving Kandinsky S2V 1.0 Lite on the $80$-item Russian set. Percentages are reported as whole numbers and sum to $100$\% for each criterion.}
\label{fig:sbs_lite}
\end{figure}

Popular metrics for talking-avatar evaluation (e.g., Sync-C) often diverge from human perception, as models can exploit them by exaggerating lip motion. Such metrics also fail to capture factors like the naturalness of articulation and avatar motion. To bridge this gap, we complement them with a side-by-side (SBS) human evaluation.

The evaluation uses two sets, in which each reference image is paired with a distinct audio clip of $5$ to $50$ seconds. The Russian-language set contains $80$ cases. The multilingual set contains about $140$ cases, roughly $30$\% English, $30$\% Russian, $10$\% Chinese, and $30$\% spread over other languages, including rare and low-resource ones. Neither set overlaps the training data. The reference images and audio are taken from open datasets and large video platforms or generated, are collected outside the training pipeline, and pass the same exclusion checks as the benchmarks (Section~\ref{subsec:provenance}). The items are selected to cover a wide range of inputs. The reference images vary in body pose, background, gender, age, and shooting conditions. The audio clips vary in dialect, speaking style, emotion, loudness, and pauses, and some contain background noise or music. Some of the references resemble amateur photos and recordings rather than studio footage. Pro is evaluated on both sets, and Lite on the Russian set.

In these comparisons, both systems receive the same reference image and speech track, and their outputs are shown in randomized left-right order without model identities. The seven criteria are speech-video synchronization, motion naturalness, hand generation, teeth and mouth generation, identity preservation, camera stability, and visual quality. For each criterion a rater selects one of four options: the left model is better, the right model is better, both are good, or neither is acceptable. Under camera stability, raters prefer the video with less zoom, drift, or shake that the prompt did not ask for. Each item is annotated by three of $20$ experienced annotators instructed on the criteria beforehand, and items are distributed uniformly across the pool, so that no single rater annotates a large share of the set.

We evaluate Pro at HD against the commercial services, which return HD output only, and run each pair at a matched resolution. We compare against LongCat 1.5 \cite{longcatteam2026longcatavatar15}, the strongest open-source baseline of Section~\ref{sec:auto_eval},\footnote{Generated with the inference code and weights released at \url{https://github.com/meituan-longcat/LongCat-Video}.} and against the closed commercial services HeyGen Avatar~IV \cite{heygen2025avatariv}\footnote{Generated through the public web interface at \url{https://app.heygen.com/home?page=photo-to-video}.} and Kling Avatar~2 \cite{klingai2026avatar2}.\footnote{Generated through the public web interface at \url{https://kling.ai/app/ai-human/new}.}

Figure~\ref{fig:sbs} reports the Russian set. Speech-video synchronization is the clearest advantage of Kandinsky S2V 1.0 Pro in all three comparisons, with $54$\% to $62$\% of judgments in its favor against $4$\% to $15$\% for the other system. Pro is also preferred for motion naturalness and hand generation, while identity preservation and camera stability are mostly ties. The only clear loss is visual quality against LongCat 1.5 at SD. At HD, Pro is slightly ahead of both commercial services on this criterion.

On the multilingual set, where Pro meets the two strongest systems of the Russian round at HD, it keeps the lead in speech-video synchronization and hand generation (Figure~\ref{fig:sbs_intl}). Here it trails both systems in visual quality. HeyGen Avatar IV is also preferred for motion naturalness, teeth and mouth generation, and identity preservation, and motion naturalness against LongCat 1.5 is even.

Figure~\ref{fig:sbs_lite} adds two SD comparisons for Lite on the Russian set. Against LongCat 1.5, Lite is preferred for speech-video synchronization and teeth and mouth generation and loses on visual quality. Against Pro, Lite loses on visual quality and synchronization, and the other criteria are mostly ties.

\subsection{Distillation Results}\label{subsec:distill_results}

Table~\ref{tab:comparison} places both distilled variants alongside their base models and the baselines. Distilled Pro leads both aggregate measures on each benchmark. Relative to base Pro, it improves CSIM, IQA, and ASE on all three benchmarks. Sync-C is slightly lower on EMTD ($8.03$ versus $8.05$) and CelebV-HQ ($7.19$ versus $7.20$), but higher on HDTF ($8.12$ versus $8.01$); Sync-D is higher on all three. Distilled Lite improves the min-max score on each benchmark, while its Sync-C is lower and Sync-D higher than base Lite.

\begin{figure}[!htbp]
\centering
\includegraphics[width=0.78\textwidth]{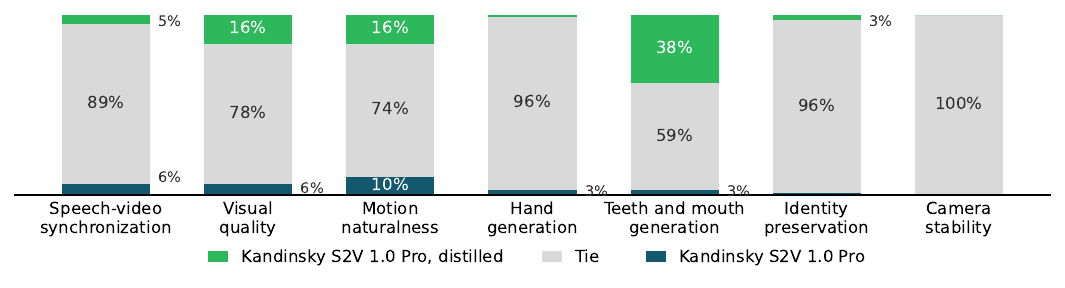}
\caption{Side-by-side human evaluation of distilled Kandinsky S2V 1.0 Pro against base Pro on the $80$-item Russian set, HD.}
\label{fig:sbs_piflow}
\end{figure}

A side-by-side human evaluation provides a complementary comparison. For Pro, we repeat the side-by-side protocol (Section~\ref{sec:human_eval}) on the $80$-item Russian set in high definition, with base Pro, the teacher, in place of a competing system. Figure~\ref{fig:sbs_piflow} reports the result. On six of the seven criteria most judgments are ties, between $74$\% and $100$\%, and in speech-video synchronization, hand generation, identity preservation, and camera stability neither model is preferred in more than $6$\% of judgments. Distilled Pro is clearly preferred in teeth and mouth generation, in $38$\% of judgments against $3$\%, and leans ahead in visual quality, $16$\% against $6$\%, and in motion naturalness, $16$\% against $10$\%. Distilled Pro is therefore not worse than base Pro on any criterion and is noticeably better in the mouth region, the hardest region for all evaluated systems (Section~\ref{sec:human_eval}).

\begin{figure}[!htbp]
\centering
\includegraphics[width=0.78\textwidth]{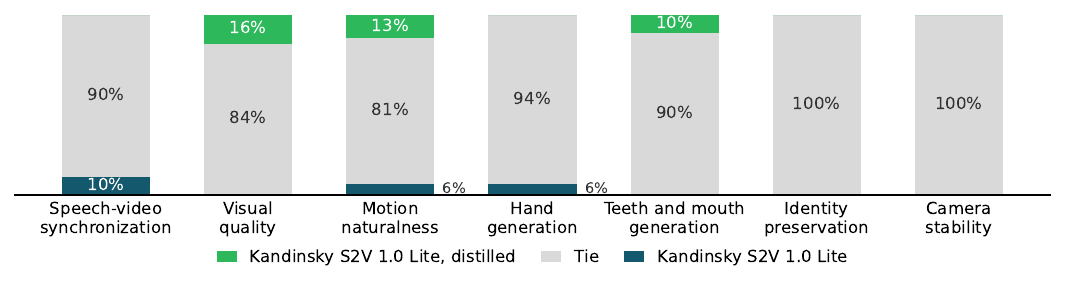}
\caption{Side-by-side human evaluation of distilled Kandinsky S2V 1.0 Lite against base Lite on the Russian set, SD.}
\label{fig:sbs_piflow_lite}
\end{figure}

For Lite we run the same comparison on the Russian set at SD, the resolution of the other Lite comparisons. Figure~\ref{fig:sbs_piflow_lite} reports the result. Most judgments are again ties, between $81$\% and $100$\%, and identity preservation and camera stability are full ties. The distilled Lite is preferred in visual quality, $16$\% against $0$\%, in motion naturalness, $13$\% against $6$\%, and in teeth and mouth generation, $10$\% against $0$\%. Base Lite keeps an edge in speech-video synchronization, $10$\% against $0$\%, and in hand generation, $6$\% against $0$\%. Together with the lower Sync-C and higher Sync-D of distilled Lite in Table~\ref{tab:comparison}, this points to a small loss of lip synchronization after distillation, larger than for Pro. A plausible reason is the step budget: $16$ policy segments may be too coarse for the 2B student to follow the guided teacher trajectory as closely as the 21B student, and a finer schedule (e.g., $24$ steps) may close the gap at moderate speed cost.

The gains after distillation, in identity for Pro and in visual quality for both models, may have two causes. First, the teacher at inference is not the continuous guided flow but one fixed $50$-step Euler discretization of it, and guidance makes that flow stiffer. The student is trained to reproduce teacher steps whose boundaries are drawn at random on every iteration, so it fits an average over many discretizations rather than the one grid the teacher is sampled on, and its policy is integrated in $128$ substeps over the trajectory rather than in $50$. Second, the policy is a smooth low-parametric function of time inside each segment, so it averages out high-frequency components of the guided velocity, which for video appear as small artifacts and flicker. Both explanations remain hypotheses; these evaluations do not isolate their effects.

\section[Conclusion]{Conclusion}\label{sec:conclusion}

We presented Kandinsky S2V 1.0 Lite ($2$B) and Kandinsky S2V 1.0 Pro ($21$B), two open models that generate long-form avatar video at $25$ FPS from a reference image, speech, and a text prompt. They obtain text and audio features from separate components of a single Qwen2.5-Omni model and use a reference-aware memory to carry visual context across chunks at three scales. Both were trained on $1.17$M single-speaker clips captioned in English and Russian. The Kandinsky S2V 1.0 family ranks among the leading models by mean rank and min-max score on EMTD, CelebV-HQ, and HDTF, with distilled Kandinsky S2V 1.0 Pro leading both aggregates on all three. In side-by-side human evaluations, Kandinsky S2V 1.0 Pro performs comparably to or better than the strongest open-source baseline and two commercial models overall and is preferred for lip synchronization. Distilled Kandinsky S2V 1.0 Pro retains strong benchmark performance with $16$ sampling steps and a $5.4\times$ end-to-end speedup over base Pro.

\subsection*{Ethical Considerations}

Kandinsky S2V 1.0 is intended to support research, education, and creative work, including content production, prototyping, and accessible tools for visual storytelling. We believe that openly available generative models should be developed and used responsibly, with respect for human rights, personal dignity, and applicable law.

Models that produce realistic videos of individuals from a reference image and speech audio introduce specific risks. They can be misused to create misleading or deceptive content, to impersonate real individuals, to spread disinformation, or to facilitate fraud and harassment. We do not support and explicitly discourage any such use. In particular, the model should not be used to depict real people without their consent, to imitate a person's appearance for deceptive purposes, to generate content that sexualizes minors or depicts non-consensual intimate imagery, to promote violence, hatred, or discrimination, or to infringe intellectual property rights.

We encourage users and downstream developers to clearly label generated content as synthetic, to comply with the laws and regulations of their jurisdictions, including those governing personal data, image and voice rights, and the disclosure of AI-generated media, and to implement appropriate safeguards, such as content moderation, when deploying the models in public-facing products.

Like other models trained on large-scale data, Kandinsky S2V 1.0 may reflect biases present in its training data, including uneven representation of demographic groups, cultures, and languages. Its outputs may also contain factual, physical, or cultural inaccuracies and should not be treated as depictions of real events. We encourage the research community to study these limitations and to report problematic behavior.
\newpage
\section[Contributors and Acknowledgments]{Contributors and Acknowledgments}\label{sec:contributors}

\subsection*{Contributors}

\textbf{Core Contributors:}
Nikolay Semenov$^{*}$, Artem Riabinin$^{*}$, Sergey Bassenko

\paragraph{Contributors:} Anna Averchenkova, Julia Agafonova, Anastasia Aliaskina, Kirill Kozlov, Lev Novitskiy

\paragraph{Project Supervisors:} Nikita Samsonov, Nikolai Gerasimenko, Denis Dimitrov
\\

\noindent{\small $^*$main and equal contributors}


\subsection*{Acknowledgments}

We thank Ivan Kirillov, Alexey Letunovskiy and Viacheslav Vasilev for their contributions.

\bibliographystyle{unsrt}
\bibliography{references}

\end{document}